\documentclass[11pt]{article}
\usepackage[table]{xcolor}
\usepackage[final]{acl}

\usepackage{times}
\usepackage{latexsym}

\usepackage[T1]{fontenc}

\usepackage[utf8]{inputenc}

\usepackage{microtype}

\usepackage{inconsolata}

\usepackage{graphicx}
\usepackage{amsmath}
\usepackage{latexsym}
\usepackage{tabularx} 
\usepackage{booktabs}
\usepackage{multirow}
\usepackage{amsmath}
\usepackage{amssymb}
\usepackage{array}
\usepackage{bm}
\usepackage{xparse}
\usepackage{mathtools}
\usepackage{pifont}
\usepackage{enumitem}
\usepackage{tcolorbox}
\tcbuselibrary{breakable}
\usepackage{listings}

\usepackage{arydshln}
\usepackage{hyperref}
\usepackage{algorithm}

\usepackage{colortbl}
\usepackage{booktabs}
\usepackage{siunitx}

\usepackage{graphicx}  
\usepackage{float}  
\usepackage{algpseudocode}
\usepackage{xcolor}
\usepackage{amsmath}
\usepackage{amssymb}
\usepackage[italicComments=true, indLines=true]{algpseudocodex} 
\usepackage{xcolor}
\definecolor{darkred}{RGB}{160, 30, 30}
\definecolor{graycomment}{RGB}{120, 120, 120}
\usepackage{inputenc}

\usepackage{arydshln}
\usepackage{hyperref}
\usepackage{algorithm}
\usepackage{colortbl}
\usepackage{booktabs}
\usepackage{siunitx}
\usepackage{algpseudocode}
\usepackage{xcolor}
\usepackage{amsmath}
\usepackage{amssymb}
\usepackage[italicComments=true, indLines=true]{algpseudocodex} 
\usepackage[table]{xcolor}
\usepackage{inputenc}
\newcommand{\chg}[1]{%
  \ifdim #1 pt > 0 pt
    {\textcolor{green!60!black}{\scriptsize (+#1)}}%
  \else\ifdim #1 pt < 0 pt
    {\textcolor{blue!70!black}{\scriptsize (#1)}}%
  \else
    {\textcolor{gray}{\scriptsize (0)}}%
  \fi\fi
}
\usepackage{subcaption}
\usepackage{placeins}

\algrenewcommand{\algorithmiccomment}[1]{\hfill \textit{\textcolor{graycomment}{// #1}}}
\title{Why Multi-Step Tool-Use Reinforcement Learning Collapses and \\How Supervisory Signals Fix It}

\author{
 \textbf{Yupu Hao\textsuperscript{1,2}},
 \textbf{Zhuoran Jin\textsuperscript{1,2}},
 \textbf{Huanxuan Liao\textsuperscript{1,2}},
 \textbf{Kang Liu\textsuperscript{1,2}},
 \textbf{Jun Zhao\textsuperscript{1,2}\thanks{Corresponding Author}}
\\
 \textsuperscript{1}The Key Laboratory of Cognition and Decision Intelligence for Complex Systems, \\Institute of Automation, Chinese Academy of Sciences, Beijing, China \\
 \textsuperscript{2}School of Artificial Intelligence, University of Chinese Academy of Sciences, Beijing, China
\\
\{haoyupu2023, liaohuanxuan2023\}@ia.ac.cn, \{zhuoran.jin, kliu, jzhao\}@nlpr.ia.ac.cn
}

\begin{document}
\maketitle
\begin{abstract}
Tool use enables large language models (LLMs) to perform complex tasks, and recent agentic reinforcement learning (RL) methods show promise for enhancing model capabilities. However, RL alone often leads to instability or limited gains in tool-use tasks. In our experiments, some models exhibit catastrophic collapse, where performance abruptly drops and tool-invocation structures fail. The analysis reveals that these failures stem from unexpected probability spikes in specific control tokens, disrupting structured execution, yet the underlying tool-use capability remains intact, merely obscured by specific formats. To address this, we systematically investigate a diverse set of supervisory signals, including off-policy supervision, hint-based guidance, erroneous example supervision, and others, applied under both synchronous and interleaved training schemes. We find that interleaving supervised fine-tuning (SFT) with RL substantially improves stability, but exhibits degraded performance under format and content out-of-distribution (OOD) evaluation. We also analyze the impact of learning rates and generalization across settings. These results highlight the importance of understanding RL failures and demonstrate how diverse supervisory signals can guide exploratory learning, enabling robust training of LLMs for complex, multi-step tool-use tasks. Our Code is available at \url{https://github.com/hypasd-art/Tool-RL-Box}.
\end{abstract}

\section{Introduction}

\begin{figure*}[t]
    \centering
    \includegraphics[width=1\textwidth]{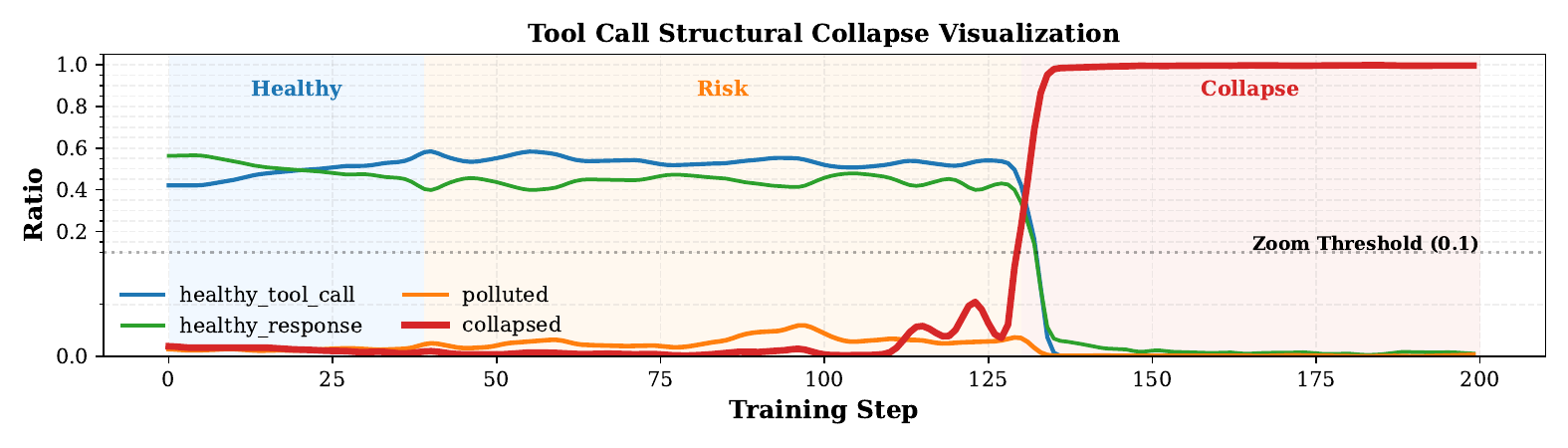}
    \caption{The changes of token appearing frequency in the RL training process of model Qwen2.5-1.5B-Instruct. Polluted indicates that irrelevant special tokens appear in the output, whereas Collapsed indicates that the output is completely invalid or incorrect.}
    \label{fig:token_probility}
    \vskip -0.2in
\end{figure*}

Tool use plays an increasingly important role in large language models (LLMs) \cite{DBLP:conf/emnlp/LiZ000YLHL23, DBLP:journals/tmlr/MialonDLNPRRSDC23, li2026agentic}. By leveraging external tools, LLMs can function as intelligent agents capable of autonomously performing complex interactive tasks \cite{DBLP:journals/fcsc/QuDWCWYXW25, DBLP:journals/corr/abs-2406-12045}. However, the multi-step and structured nature of tool-based interactions, together with the diversity of tool feedback, introduces substantial challenges for improving model capabilities. To mitigate these issues, prior works have focused on improving tool-use performance through optimized interaction frameworks \cite{DBLP:conf/iclr/QinLYZYLLCTQZHT24}, large-scale synthesis of high-quality trajectories \cite{DBLP:conf/acl/HaoCJL0LZ25, DBLP:conf/iclr/Liu0ZHYL0GLY0WN25, prabhakar2025apigen}, and refined supervised or unsupervised training methodologies \cite{DBLP:conf/iclr/Liu0ZHYL0GLY0WN25, DBLP:journals/corr/abs-2510-10197}. Particularly recent advances in agentic RL \cite{DBLP:journals/corr/abs-2510-10197, wang2025ragen, mai2025agent} have motivated growing interest in RL as a more principled framework for agentic interaction, as agentic RL has demonstrated strong performance across a variety of complex tasks \cite{zhang2025landscape, wu2025agentic}. 

However, in practice, RL-based optimization often suffers from \textbf{training instability} and \textbf{limited performance gains} in tool-use settings. As shown in Figure~\ref{fig:token_probility}, erroneous response frequencies can rise sharply at certain stages. In our experiments, we observe a surprising failure mode where model performance can abruptly collapse to near zero, accompanied by a breakdown of valid tool-invocation structures. Through detailed analysis, we find that these failures are not caused by a loss of reasoning ability, but are instead driven by unexpected probability amplification of specific control tokens, which gradually distorts structured generation into degenerate execution patterns.

These observations indicate that RL alone is insufficient for stable, long-horizon structured generation. We find supervised fine-tuning on high-quality tool-use trajectories provides a strong initialization, improving early performance and preventing collapse. Nonetheless, prior work suggests that models trained with SFT can be limited in terms of generalization and performance compared to RL \cite{chu2025sft}. Therefore, we aim to \textbf{investigate how different supervisory signals interact with RL optimization and whether supervision can fundamentally stabilize multi-turn agentic training, attempting to combine them to mitigate their shortcomings}. Yet, a key open question remains: \textit{how should supervisory signals be integrated into RL training, and which forms of supervision are most effective for enhancing stability?} 

To address this issue, we systematically investigate a broad family of supervisory signals, including SFT then RL, off-policy supervision, and erroneous trajectory supervision, under both synchronous and interleaved training paradigms. Additionally, we introduce Hint-based guidance by prepending correct hints before the model generates responses to help generation, and remove these hints during optimization. We also propose Process Reflection Supervision, which reflections are extracted from intermediate reasoning steps in the training process for the summarization the strengths and weaknesses of each steps alongside original error trajectories to construct the SFT data, further improves both stability and final performance. Our results show that simply mixing supervision with RL is insufficient: synchronous methods often suffer from distribution mismatch, while interleaved training with supervised fine-tuning (SFT) significantly improves stability but may degrade performance under format- and content-level out-of-distribution (OOD) evaluation.

We further analyze key factors affecting stability and generalization, including learning rate sensitivity and cross-setting transfer behavior. Across these studies, we find that supervisory signals play a central role in controlling structural behavior during RL, beyond what scalar rewards alone can achieve.

Overall, our work highlights that agentic RL failure is primarily a \emph{structural collapse problem} rather than a capability limitation, and demonstrates that carefully designed supervisory signals can effectively regulate token-level execution patterns, enabling more stable and robust tool-use learning in LLMs.

Our work makes three key contributions:

\begin{itemize}

    \item \textbf{We identify structural collapse as a fundamental failure mode in multi-turn agentic RL.}
    We show that RL on tool-use tasks does not primarily degrade reasoning ability, but induces a \emph{structural collapse} in which generation degenerates into malformed control-token sequences, breaking tool-invocation structure while preserving underlying task competence.
    
    \item \textbf{We uncover a token-level mechanism underlying RL instability.}
    We find that RL disproportionately amplifies special control tokens, leading to a redistribution of policy mass. This reveals that instability arises from \emph{control-token dynamics} rather than capability degradation, providing a mechanistic explanation for collapse.
    
    \item \textbf{We conduct a systematic study of supervisory signals for stabilizing multi-turn agentic RL.}
    We systematically explore six supervisory configurations across two model families and categorize them into synchronous and interleaved paradigms. We further introduce Process Reflection Supervision, which converts intermediate trajectory information into textual supervisory signals. Empirically, interleaved training consistently improves stability and generalization, while synchronous methods are prone to distribution mismatch.

\end{itemize}

\begin{figure*}[t]
    \centering
    \begin{minipage}[b]{0.48\textwidth}
        \centering
        \includegraphics[width=\linewidth]{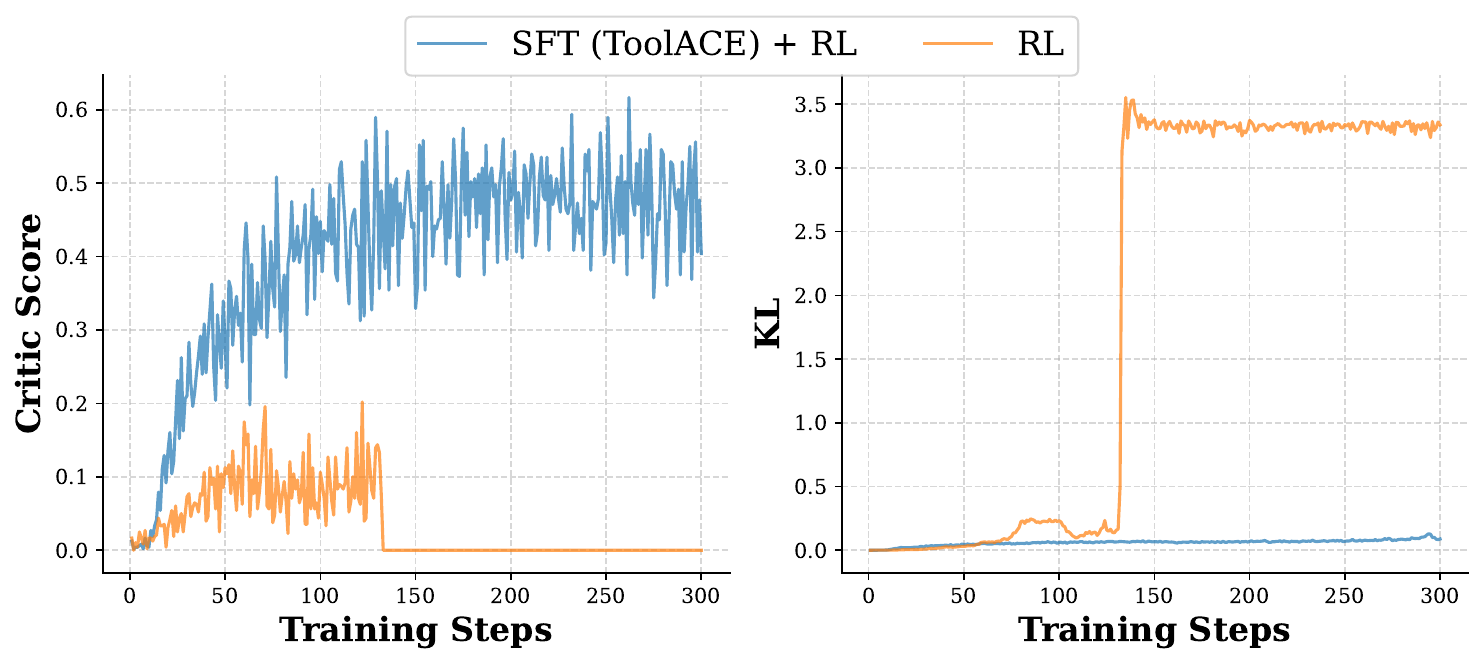}
    \end{minipage}
    \hfill
    \begin{minipage}[b]{0.48\textwidth}
        \centering
        \includegraphics[width=\linewidth]{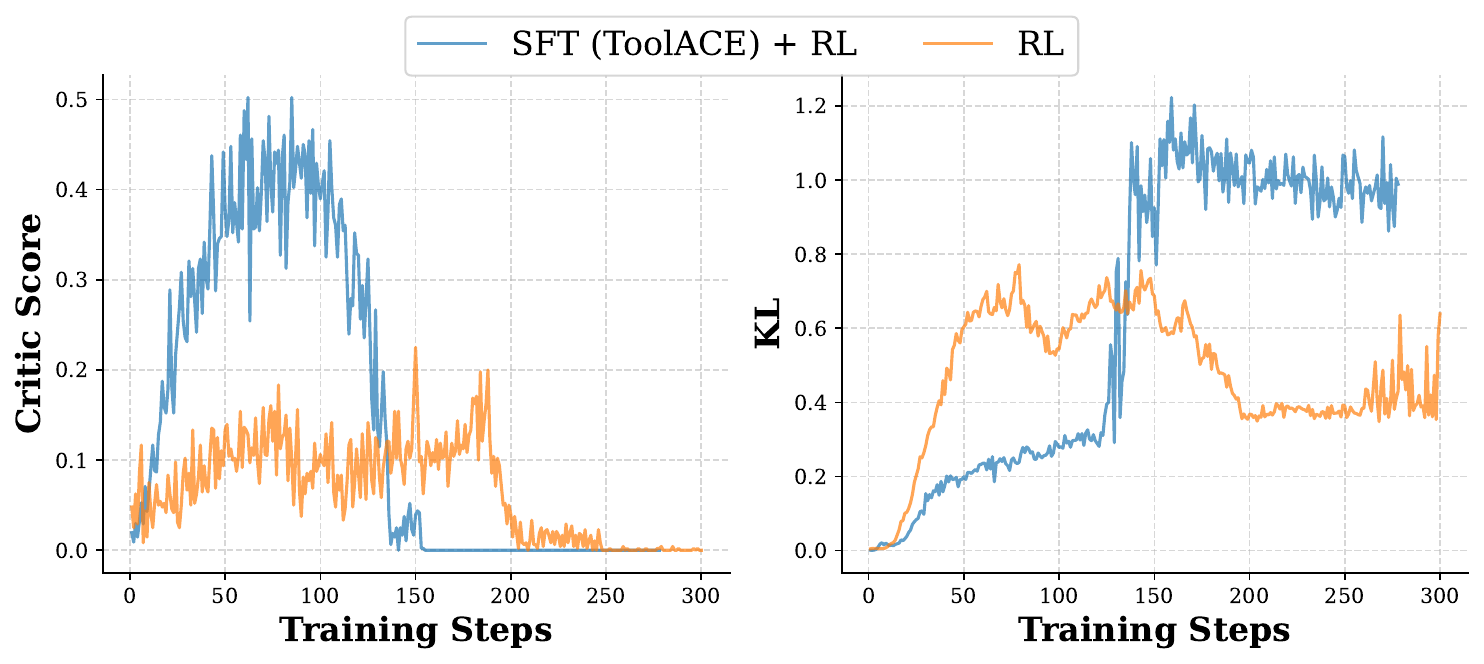}
    \end{minipage}

    \caption{The training dynamics during the training process on the BFCL-V3 dataset. \textbf{Left}: Qwen2.5-1.5B-Instruct. \textbf{Right:} Qwen3-1.7B.}
    \label{fig:curves_training}
    \vskip -0.2in
\end{figure*}

\section{Related Works}
\subsection{Tool Learning}

Recent advances have extended LLMs with tool-use capabilities \cite{DBLP:conf/aaai/HaoCJL0LZ25, DBLP:journals/corr/abs-2504-03601, DBLP:conf/iclr/QinLYZYLLCTQZHT24}, enabling interaction with external APIs and environments beyond text generation \cite{DBLP:journals/corr/abs-2510-10197}. Through tool invocation, models can perform task execution \cite{DBLP:conf/emnlp/LiZ000YLHL23, DBLP:conf/acl/TrivediKHMDLGSB24}, information retrieval \cite{DBLP:journals/corr/abs-2503-09516}, and planning \cite{xie2024travelplanner}, improving their ability to solve complex tasks efficiently. Prior works have enhanced tool-use capabilities through reasoning frameworks \cite{DBLP:conf/iclr/QinLYZYLLCTQZHT24, DBLP:conf/iclr/YaoZYDSN023}, supervised fine-tuning \cite{DBLP:journals/corr/abs-2510-10197, DBLP:journals/corr/abs-2505-00024, DBLP:journals/corr/abs-2503-23383}, and reinforcement learning \cite{DBLP:journals/corr/abs-2501-03262, DBLP:journals/corr/abs-2402-03300}. Although RL improves tool-invocation performance \cite{DBLP:journals/corr/abs-2504-13958, DBLP:journals/corr/abs-2509-01055, DBLP:journals/corr/abs-2509-02479}, its effectiveness strongly depends on the base model’s prior tool knowledge. Smaller or weakly initialized models often fail to benefit from RL, while high-quality SFT data alone can already provide strong performance. This motivates incorporating supervisory signals into RL to stabilize multi-turn interactions and improve tool-use learning.

\subsection{Expert Trajectories in Reinforcement Learning}

RL improves models through trajectory exploration and reward-based updates \cite{DBLP:journals/corr/abs-2402-03300, DBLP:journals/corr/abs-2507-04136}, but can stagnate when sampling quality is poor. Recent works mitigate this by incorporating expert or ground-truth trajectories into RL \cite{fu2025srft, DBLP:journals/corr/abs-2505-16984}. For instance, LUFFY \cite{DBLP:journals/corr/abs-2504-14945} replaces part of sampled trajectories with expert ones and reweights low-probability tokens, while ReLIFT \cite{ma2025learning} alternates RL with targeted SFT. Other methods use partial correct answers \cite{zhang2025adhint, zhang2025stephint, DBLP:journals/corr/abs-2503-23905, DBLP:journals/corr/abs-2510-02245} or retrieved experiences \cite{DBLP:journals/corr/abs-2507-23361, DBLP:journals/corr/abs-2508-06433}. However, these methods are mainly studied in single-turn reasoning tasks (e.g., mathematics), and their effectiveness in multi-turn tool-use settings remains underexplored.

\section{Preliminary}

We represent multi-turn tool utilization trajectory $\tau$ by LLMs as a sequence of actions and feedbacks: $\tau = (a_1, r_1, a_2, r_2, \dots, a_T, r_T)$. Each action $a_t \in \mathcal{A}$ is drawn from an action space $\mathcal{A}$ that includes both \textbf{feasible tool invocations} and \textbf{natural language responses}. At each step, the feedback $r_t \in \mathcal{R}$ is given based on current actionand the preceding trajectory: 


\begin{equation}
\footnotesize
    r_t = R(a_t \mid a_1, r_1, \dots, a_{t-1}, r_{t-1}) \in \mathcal{R}
\end{equation}

where $\mathcal{R}$ denotes the feedback space, encompassing both \textbf{environment responses} (e.g., tool execution outputs) and \textbf{user responses}. 

Conditioned on the accumulated interaction history, the model samples the next action: 

\begin{equation}
\footnotesize
    a_t \sim P(a_t \mid a_1, r_1, \dots, a_{t-1}, r_{t-1})
\end{equation}

\section{Method}
In this section, we first analyze the \emph{training instability and limited performance gains} observed under pure reinforcement learning settings. By examining model-generated trajectories during training, we identify the underlying causes of these failures. In contrast, introducing supervised fine-tuning before RL substantially improves stability and performance, highlighting the importance of supervisory signals. Motivated by this observation, we further conduct a systematic study of different supervisory signals and integration strategies in agentic RL, analyzing their effects on stability, structural behavior, and generalization.

\subsection{Experiment Setup}

\textbf{Dataset.} We conduct experiments on BFCL-V3 \cite{DBLP:conf/icml/PatilMYJSSG25}, a multi-turn tool-use benchmark where models must invoke multiple tools across interactive environments to answer user queries and follow-up questions. We focus on four challenging settings: \textit{Base}, \textit{Miss Func}, \textit{Miss Param}, and \textit{Long Context}. For training, we randomly sample 100 instances from each of the first three settings, resulting in 300 questions following previous work \cite{DBLP:journals/corr/abs-2510-10197}, which proves the performance can be improved by small amount of data, while reserving the remaining data for evaluation. Due to excessive context length, \textit{Long Context} is excluded from training. We also augment supervised fine-tuning with converted ToolACE \cite{DBLP:conf/iclr/Liu0ZHYL0GLY0WN25} data to study the impact of training distribution. More details please refer to Appendix~\ref{appendix:training_details}


\textbf{Model.} For our analysis experiments, we use Qwen2.5-1.5B-Instruct \cite{team2024qwen2} and Qwen3-1.7B \cite{DBLP:journals/corr/abs-2505-09388} as base models. During the SFT stage, we decompose each multi-turn interaction into multiple single-turn instances and train the model on these isolated steps. In contrast, during RL, we adopt a full multi-turn tool-use setting, where the model must complete the entire interaction trajectory, including all tool invocations, to receive positive feedback. For Qwen3-1.7B, we use the non-thinking variant. Further implementation details and reward design are provided in Appendix~\ref{appendix:training_details}.

\begin{figure*}[tp]
    \centering
    \includegraphics[width=0.9\textwidth]{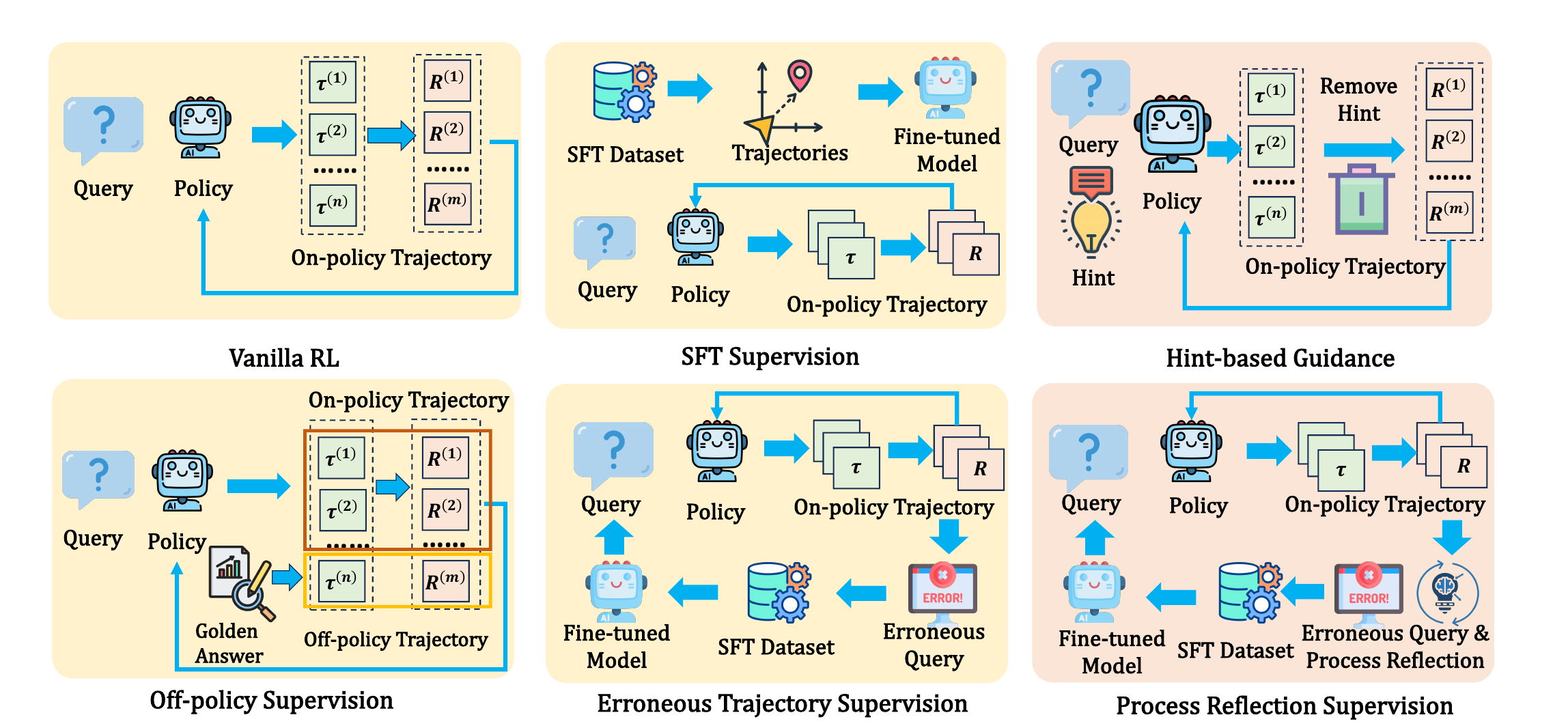}
    \caption{The training frameworks of different methods, where the Hint-based Guidance and Process Reflection Supervision is proposed by our.}
    \label{fig:model}
    \vskip -0.2in
\end{figure*}

\subsection{Problem of catastrophic collapse in the RL}
As shown in Figure~\ref{fig:curves_training}, direct RL on Qwen2.5-1.5B-Instruct often causes \textbf{catastrophic collapse}, with sudden reward drops and sustained KL divergence spikes. Qwen3-1.7B exhibits milder instability, but pure RL gains remain limited and may degrade over time. Applying SFT on an additional tool-use dataset before RL stabilizes training: for Qwen2.5-1.5B-Instruct, reward and KL curves evolve smoothly; for Qwen3-1.7B, RL may still occasionally collapse after SFT, but achieves substantially higher performance prior to instability.

\subsection{Analysis of the reasons of collapse}
\label{sec:analysis}

By jointly analyzing model-generated trajectories throughout RL training, we identify a consistent pattern of \emph{structural collapse}. Importantly, we find that the observed collapse does not primarily reflect a degradation of the model’s underlying capability. Instead, the collapse progressively shifts generation toward malformed control-token patterns that corrupt the structural integrity of tool invocation. Specifically, special control tokens used in tool-calling formats, such as \texttt{<tool\_call>} and \texttt{<|im\_end|>}, become increasingly confused during training. As RL optimization proceeds, these tokens are over-amplified and incorrectly combined as shown in Tabel~\ref{tab:instance}, causing the model to drift from valid tool-use trajectories toward degenerate termination patterns under particular prompt formats.

To systematically characterize this phenomenon, we categorize model outputs into four structural states. \textbf{Healthy Tool Call} denotes well-formed tool invocations that strictly follow the required schema. \textbf{Healthy Response} corresponds to natural-language responses and contains no tool-related artifacts. \textbf{Text Pollution} represents an intermediate failure stage where incomplete tool tags or misplaced control tokens begin to appear. Finally, \textbf{Collapsed} denotes severe degeneration, where generation converges to meaningless minimal termination sequences such as \texttt{<tool\_call><|im\_end|>} with almost no semantic content.

As shown in Figure~\ref{fig:token_probility}, Qwen2.5-1.5B-Instruct initially maintains a clear separation between \textbf{tool-call} and \textbf{text-response} generation pathways, producing structurally valid outputs during early training. However, as RL progresses, fragmented control-token patterns increasingly appear within natural-language responses (see Table~\ref{tab:instance}), indicating progressive corruption of structural boundaries. In later stages, the probability mass becomes dominated by minimal termination sequences \texttt{<tool\_call><|im\_end|>}, leading to complete structural collapse. Detailed definitions and additional results are provided in Appendix~\ref{appendix:analysis}.

These results suggest that instability in multi-turn tool-use RL arises from insufficient structural supervision during exploration. Without constraints that preserve tool-invocation structure, RL disproportionately reinforces certain control tokens, gradually redistributing policy mass toward degenerate execution patterns. Interestingly, results in Table~\ref{tab:merged_ood_results} further show that direct GRPO training remains relatively stable under format changes in OOD evaluation. This indicates that the underlying tool-use capability is not fully lost during collapse; instead, performance is highly sensitive to control-token distributions that can either expose or temporarily mask the learned capability.

Overall, our analysis reveals that tool-use RL is fundamentally more sensitive to structural token dynamics than standard reasoning tasks. Without proper supervision or constraints, these tokens can be easily conflated, leading to structural contamination and, ultimately, training collapse.

\subsection{How to improve the collapse problem}

Unlike prior works studying isolated supervision strategies, we propose a unified framework to analyze how supervisory signals interact with multi-turn agentic RL. We organize methods into synchronous and interleaved paradigms, enabling a controlled comparison of their effects on stability, structural collapse, and generalization in tool-use RL. Within this framework, we systematically study five supervisory signals: SFT-then-RL, off-policy supervision, hint-based guidance, erroneous trajectory supervision, and process reflection supervision. Structural diagrams are shown in Figure~\ref{fig:model}, with reward details in Appendix~\ref{appendix:training_details}.

\paragraph{SFT Supervision (SFT then RL)} 
This supervisory signal follows a staged pipeline where SFT is performed before RL. In the SFT phase, the policy $\pi_\theta$ is trained to imitate expert trajectories $\tau$ given queries $q$ from dataset $D_{\text{SFT}}$: 

\begin{equation}
\footnotesize
\mathcal{L}_{\text{SFT}}(\theta) = - \frac{1}{|D_{\text{SFT}}|} \sum_{(q, \tau) \in D_{\text{SFT}}} \sum_{t=1}^{|\tau|} \log \pi_\theta(a_t \mid q, \tau_{<t}),
\end{equation}
this objective encourages instruction-following and basic reasoning abilities. However, its effectiveness is constrained by dataset coverage and lacks the ability to explore behaviors beyond the provided demonstrations. 

\paragraph{Off-policy Supervision (OPS)} 
Some methods like LUFFY replace the sampled responses with distilled content or ground truth responses \cite{DBLP:journals/corr/abs-2504-14945}. Inspired by these, we introduce off-policy supervision by partially replacing model-generated rollouts with ground-truth trajectories, thereby strengthening guidance from correct behaviors. Specifically, we mix high-quality SFT trajectories with on-policy rollouts. For each trajectory $\tau_i$, the normalized advantage is computed as: 
\begin{equation}
\footnotesize
\hat{A}_i = \frac{R(\tau_i) - \mathrm{mean}(G_\text{on} \cup G_\text{off})}
{\mathrm{std}(G_\text{on} \cup G_\text{off})},
\end{equation}
where $G_\text{on}$ and $G_\text{off}$ denote the reward sets from on-policy and off-policy trajectories, respectively. Unlike LUFFY \cite{DBLP:journals/corr/abs-2504-14945}, which modifies the importance ratio for off-policy data, we apply the same policy ratio to all trajectories $r_{k,t}(\theta) = \pi_\theta(\tau_{k,t} \mid q, \tau_{k,<t})/\pi_{\theta_\text{old}}(\tau_{k,t} \mid q, \tau_{k,<t})$ to fairly compare the performance of different supervisory signals without other factors. This yields the off-policy supervision surrogate objective:
\begin{equation}
\footnotesize
\mathcal{J}_\text{OPS}(\theta) = \frac{1}{Z} \sum_{\tau_i \in \mathcal{D}_{\text{on}} \cup \mathcal{D}_{\text{off}}} \sum_{t=1}^{|\tau_i|} 
\mathrm{CLIP}\big(r_{k,t}(\theta), \hat{A}_i, \epsilon \big),
\end{equation}
where $Z = \sum_i |\tau_i|$.
This formulation treats on-policy and off-policy trajectories uniformly, enabling high-quality demonstrations to guide learning while preserving on-policy exploration. 

\paragraph{Hint-based Guidance (HBG)}

To guide the model towards correct trajectories, we prepend a hint $\mathcal{H}_q$ to the query $q$ during sampling. The hint provides a textual description of how to approach or solve the task, and can be obtained either through manual annotation or generated automatically by a large language model. At each timestep, the trajectory is generated token by token as: 

\begin{equation}
\footnotesize
\tau_{i,t} \sim \pi_\theta(\tau_t \mid q, \tau_{i,<t}, \mathcal{H}_q),
\end{equation}
for policy optimization, the hint $\mathcal{H}_q$ is removed, and the sampled trajectory $\tau_i$ is treated as a standard on-policy rollout conditioned only on $q$. This design allows hints to influence exploration while ensuring that the learned policy does not depend on auxiliary guidance at inference time.

\paragraph{Erroneous Trajectory Supervision (ETS)}
\label{sec:erroneous}
Some methods adopt the interleaved training paradigm of SFT and RL in the training process.
Inspired by these \cite{ma2025learning,lai2025computerrlscalingendtoendonline}, we adopt an interleaved RL and SFT paradigm into the tool-learning tasks, constructing SFT data explicitly from RL failure cases. Starting from an initial policy $\pi_{\theta_0}$, RL collects interaction trajectories, and any query whose sampled trajectories all fail is labeled as erroneous. We then create supervised training data from these failures using the ground-truth solutions to enhance model capability.

Formally, at iteration $k$, the error-derived supervision set is defined as:
\begin{equation}
\footnotesize
\mathcal{E}_k = \bigcup_{q \in Q} \{ (q, \tau_{\text{label}}) \mid \forall \tau' \in \mathcal{T}_q,\; R(\tau') = 0 \},
\end{equation}
where $\mathcal{T}_q$ denotes the set of trajectories sampled for query $q$, and $R(\tau)$ is the reward, $\tau_{\text{label}}$ is the ground truth answer. For each trajectory, we construct supervised training pairs conditioned on the same state or prefix. The resulting supervision loss is:
\begin{equation}
\footnotesize
\mathcal{L}_{\mathrm{err}}(\theta)= - \frac{1}{|\mathcal{E}_k|} \sum_{(q, \tau) \in \mathcal{E}_k} \sum_{t=1}^{|\tau|} \log \pi_\theta(a_t \mid q, \tau_{<t}),
\end{equation}
where $a_t$ denotes the ground-truth action at step $t$.
We interleave this corrective SFT step with standard RL updates, and gradually increase the proportion of RL steps over iterations, following the scheduling strategy in ReLIFT \cite{ma2025learning}.

\begin{table*}[!ht]
    \centering
    \small
    \renewcommand{\arraystretch}{1.0}
    \setlength{\tabcolsep}{3pt}
    
    \begin{tabular}{l l r c r c r c r c r c}
        \toprule
        \textbf{Model} & \textbf{Method} & \textbf{Base} & $\Delta$ & \textbf{Miss F.} & $\Delta$ & \textbf{Miss P.} & $\Delta$ & \textbf{Long C.} & $\Delta$ & \textbf{Average} & $\Delta$ \\
        \midrule
        
        \multirow{9}{*}{\textbf{Qwen2.5-1.5B}} 
        & Vanilla 
        & 4.0 & - & 5.0 & - & 1.0 & - & 4.0 & - & 3.50 & - \\

        & $\text{SFT}_{\text{BFCL}}$
        & 15.0 & \chg{15.0}
        & 4.0 & \chg{14.0}
        & 6.0 & \chg{16.0}
        & 7.0 & \chg{8.0}
        & 16.75 & \chg{13.25} \\

        & GRPO
        & 0.0 & \chg{-4.0}
        & 0.0 & \chg{-5.0}
        & 0.0 & \chg{-1.0}
        & 0.0 & \chg{-4.0}
        & 0.0 & \chg{-3.5} \\

        \cmidrule{2-12}

        & $\text{SFT}_{\text{BFCL}}$ + RL
        & 21.0 & \chg{17.0}
        & 22.0 & \chg{17.0}
        & 19.0 & \chg{18.0}
        & 7.0 & \chg{3.0}
        & 17.25 & \chg{13.75} \\

        & $\text{SFT}_{\text{ToolACE}}$ + RL
        & 23.0 & \chg{19.0}
        & 23.0 & \chg{18.0}
        & 13.0 & \chg{12.0}
        & 10.0 & \chg{6.0}
        & 17.25 & \chg{13.75} \\

        \rowcolor[HTML]{F1F8E9} \cellcolor{white} & OPS
        & 1.0 & \chg{-3.0}
        & 3.0 & \chg{-2.0}
        & 1.0 & \chg{0.0}
        & 1.0 & \chg{-3.0}
        & 1.50 & \chg{-2.0} \\

        \rowcolor[HTML]{F1F8E9} \cellcolor{white} & HBG
        & 0.0 & \chg{-4.0}
        & 0.0 & \chg{-5.0}
        & 0.0 & \chg{-1.0}
        & 0.0 & \chg{-4.0}
        & 0.0 & \chg{-3.5} \\

        \rowcolor[HTML]{F1F8E9} \cellcolor{white} & ETS
        & 26.0 & \chg{22.0}
        & 25.0 & \chg{20.0}
        & 16.0 & \chg{15.0}
        & 13.0 & \chg{9.0}
        & 20.0 & \chg{16.5} \\

        \rowcolor[HTML]{F1F8E9} \cellcolor{white} & PRS
        & \textbf{31.0} & \chg{27.0}
        & \textbf{25.0} & \chg{20.0}
        & \textbf{26.0} & \chg{25.0}
        & \textbf{21.0} & \chg{17.0}
        & \textbf{25.75} & \chg{22.25} \\
    
        \midrule

        \multirow{10}{*}{\textbf{Qwen3-1.7B}} 
        & Vanilla 
        & 14 & - & 11 & - & 14 & - & 11 & - & 12.5 & - \\

        & $\text{SFT}_{\text{BFCL}}$
        & 23 & \chg{9}
        & 20 & \chg{9}
        & 23 & \chg{9}
        & 14 & \chg{3}
        & 20 & \chg{7.5} \\

        & $\text{SFT}_{\text{ToolACE}}$
        & 12 & \chg{-2}
        & 10 & \chg{-1}
        & 5 & \chg{-9}
        & 8 & \chg{-3}
        & 8.75 & \chg{-3.75} \\

        & GRPO
        & 2 & \chg{-13}
        & 0 & \chg{-11}
        & 2 & \chg{-13}
        & 2 & \chg{-10}
        & 1.5 & \chg{-11.0} \\

        \cmidrule{2-12}

        & $\text{SFT}_{\text{BFCL}}$ + RL
        & 0 & \chg{-14}
        & 0 & \chg{-11}
        & 0 & \chg{-14}
        & 0 & \chg{-11}
        & 0 & \chg{-12.5} \\

        & $\text{SFT}_{\text{ToolACE}}$ + RL
        & 0 & \chg{-14}
        & 0 & \chg{-11}
        & 0 & \chg{-14}
        & 0 & \chg{-11}
        & 0 & \chg{-12.5} \\

        \rowcolor[HTML]{F1F8E9} \cellcolor{white} & OPS
        & 0 & \chg{-14}
        & 0 & \chg{-11}
        & 0 & \chg{-14}
        & 0 & \chg{-11}
        & 0 & \chg{-12.5} \\

        \rowcolor[HTML]{F1F8E9} \cellcolor{white} & HBG
        & 1 & \chg{-13}
        & 0 & \chg{-11}
        & 1 & \chg{-13}
        & 1 & \chg{-10}
        & 0.75 & \chg{-11.75} \\

        \rowcolor[HTML]{F1F8E9} \cellcolor{white} & ETS
        & \textbf{26} & \chg{12}
        & \textbf{27} & \chg{14}
        & \textbf{19} & \chg{5}
        & \textbf{21} & \chg{10}
        & \textbf{23.25} & \chg{10.75} \\

        \rowcolor[HTML]{F1F8E9} \cellcolor{white} & PRS
        & 23 & \chg{9}
        & \textbf{22} & \chg{11}
        & 17 & \chg{3}
        & 16 & \chg{5}
        & 19.5 & \chg{7.0} \\

        \bottomrule
    \end{tabular}
    \caption{Experimental results on Qwen series models. Best results within each model group are highlighted in \textbf{bold}. Detailed training dynamics of individual metrics are provided in the Appendix~\ref{appendix:detailed_evaluation}.}
    
    \label{tab:qwen_results}
\end{table*}

\paragraph{Process Reflection Supervision (RPS)} 
In multi-turn agentic RL, sampled trajectories contain rich intermediate information beyond terminal success or failure. Prior work \cite{DBLP:journals/corr/abs-2509-14480} used process rewards for guidance, but scalar rewards offer only coarse supervision and often miss nuanced reasoning or tool-use errors. We propose \emph{process reflection supervision}, converting implicit trajectory information into explicit textual guidance.

Specifically, trajectories from the current policy $\pi_{\theta_k}$ are fed into an LLM (or auxiliary analyzer) to generate textual reflections analyzing intermediate decisions, structural mistakes, and reasoning gaps, denoted $r_\tau^{\mathrm{ref}}$. The resulting reflection-augmented dataset is:

\begin{equation}
\footnotesize
\mathcal{R}_{k} = \{(q, \tau, r_\tau^{\mathrm{ref}})\},
\end{equation}
to further improve the model’s adherence to correct tool-use formats, we jointly train with erroneous trajectories collected as described in Section~\ref{sec:erroneous}. The reflection loss is defined as: 
\begin{equation} 
\footnotesize
\begin{aligned}
\mathcal{L}_{\mathrm{ref}}(\theta)
& = - \frac{1}{|\mathcal{R}_k|} \sum_{(q, r_\tau^{\mathrm{ref}}) \in \mathcal{R}_k} 
\log \pi_\theta(r_\tau^{\mathrm{ref}} \mid q) \\ 
& - \frac{1}{|\mathcal{E}_k|} \sum_{(q, \tau) \in \mathcal{E}_k} \sum_{t=1}^{|\tau|} \log \pi_\theta(a_t \mid q, \tau_{<t}),
\end{aligned}
\end{equation}
the prompt and examples are in Appendix~\ref{sec:appendix_prompt_and_example}.

\section{Experiments}

\subsection{Dataset and Models}
We train the models on \textit{Qwen2.5-1.5B-Instruct} and \textit{Qwen3-1.7B} and evaluated them on BFCL-V3. To assess generalization, we further tested the models on ACEBench. Detailed training settings are provided in the Appendix~\ref{appendix:training_details}.

\subsection{Main Results} 
Based on Table~\ref{tab:qwen_results}, both Qwen2.5-1.5B-Instruct and Qwen3-1.7B show negligible baseline performance, highlighting the difficulty of multi-turn tool-use without targeted training.

\begin{figure*}[t]
    \centering
    \begin{minipage}[b]{0.48\linewidth}  
        \centering
        \includegraphics[width=\linewidth]{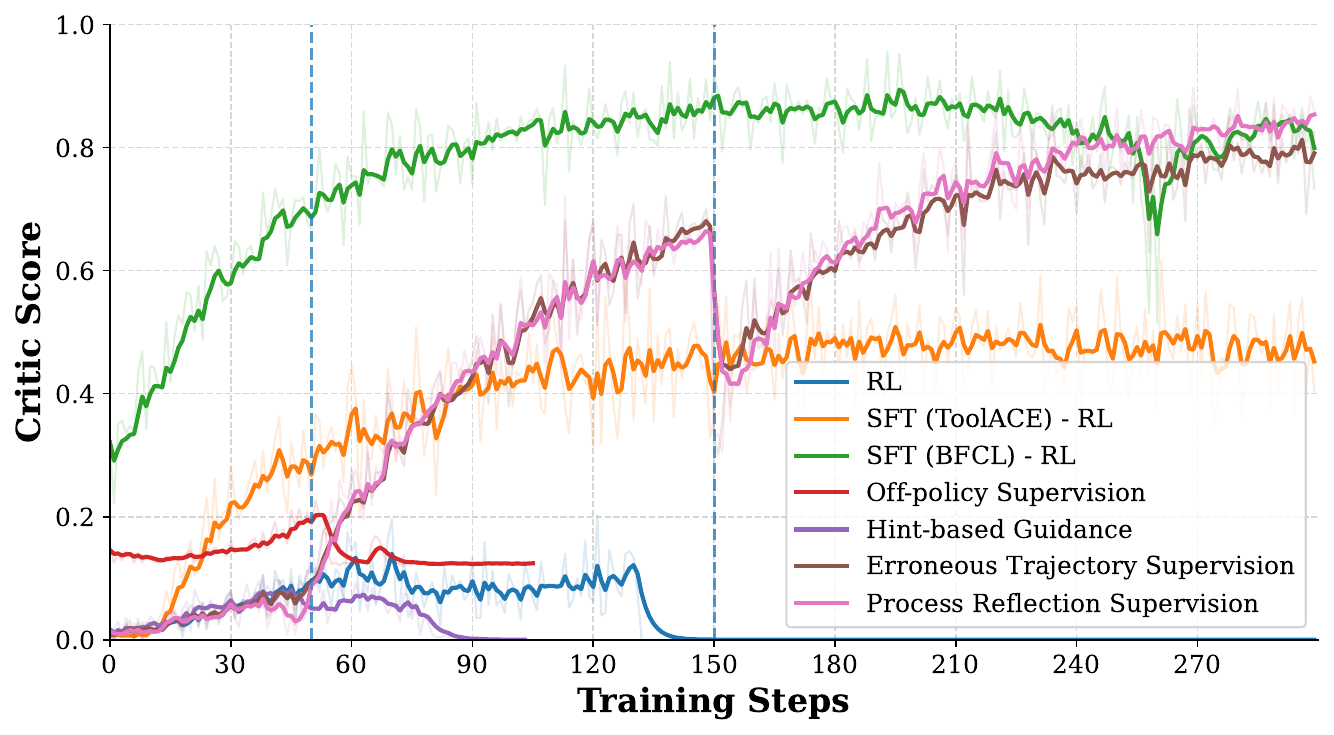}
    \end{minipage}
    \hfill
    \begin{minipage}[b]{0.48\linewidth}  
        \centering
        \includegraphics[width=\linewidth]{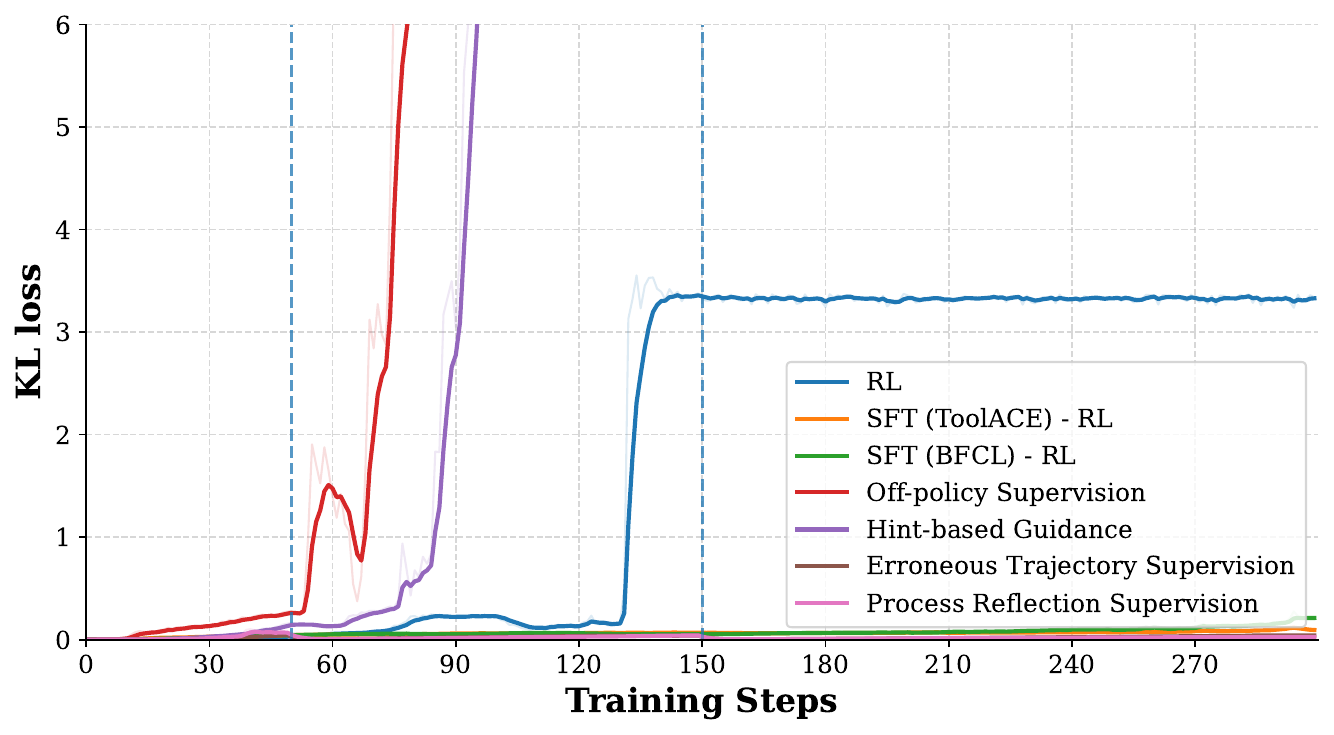}
    \end{minipage}

    \caption{The training dynamics during the training process on the BFCL-V3 dataset of Qwen2.5-1.5B-Instruct.}
    \label{fig:curves_training_con}
\end{figure*}

For Qwen2.5-1.5B-Instruct, SFT on BFCL raises the performance floor, with subsequent GRPO providing further refinement, while GRPO alone yields no meaningful gains and its stochasticity often causes catastrophic collapse, indicating that prior structural supervision is indispensable for agentic RL. Among supervisory signals, Off-Policy Supervision and Hint-Based Guidance offer limited benefits. KL divergence during synchronous training is substantially higher than in standard RL (Figure~\ref{fig:curves_training_con}), likely due to rapid output distribution shifts when trajectories are not sampled strictly on-policy, destabilizing optimization. This calls for a more granular policy-update and sampling synchronization framework. Process Reflection Supervision achieves the highest average score (25.75), followed by Erroneous Trajectory Supervision; notably, PRS's reasoning-heavy format, though diverging from standard tool-calling syntax, provides critical logical scaffolding that enhances tool-use competence. These results suggest interleaved training, decoupling SFT-like structural updates from RL exploration—is fundamentally more stable than synchronous training.

For Qwen3-1.7B, trends are consistent, but SFT prior to RL does not yield strong improvements. As shown in Figure~\ref{fig:curves_training_con3}, the model initially outperforms direct RL but eventually suffers catastrophic collapse. Based on Section~\ref{appendix:influence_qwen3_prompt}, we attribute this instability to the model's ``thinking'' mode disrupting structured trajectory learning during training sampling.

\begin{figure}[t]
    \centering
    \includegraphics[width=1\columnwidth]{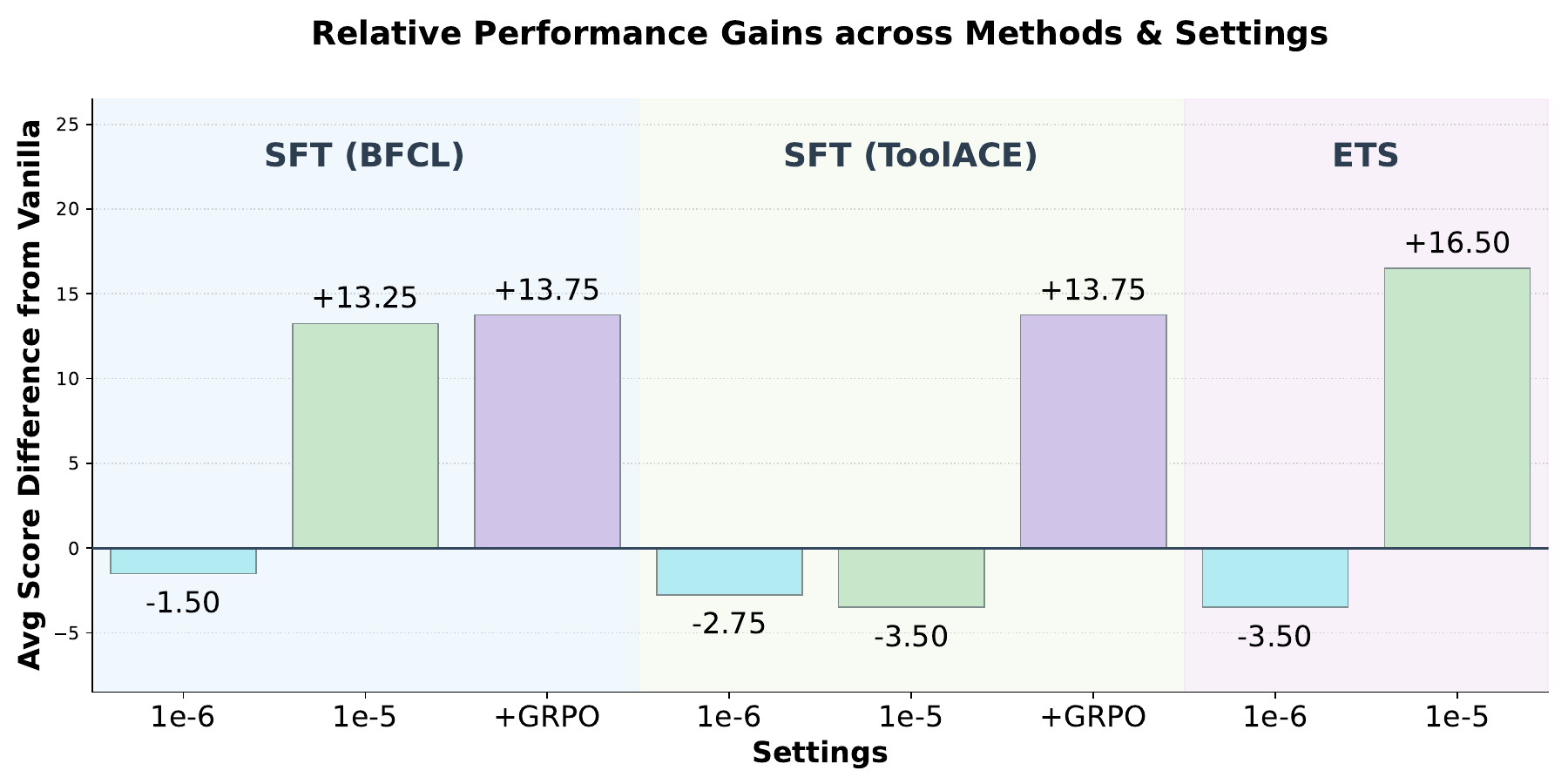}
    \caption{Results on Qwen2.5-1.5B-Instruct under different learning rate. The Y-axis illustrates performance gains and losses relative to the Vanilla baseline (0.0 line), while the X-axis represents various hyperparameter configurations and training stages, with \textit{+ GRPO} indicating RL application on models trained at a $1\times 10^{-5}$ lr.}
    \label{fig:qwen25_results}
    \vskip -0.2in
\end{figure}

\subsection{The Analysis of Learning Rate}
We evaluate the effect of learning rate on Qwen2.5-1.5B-Instruct across training configurations in Figure~\ref{fig:qwen25_results}. A conservative learning rate of $1 \times 10^{-6}$ yields decreased gains, suggesting that overly cautious updates cannot stabilize multi-turn agentic behaviors. Increasing the learning rate to $1 \times 10^{-5}$ improves SFT (BFCL), while SFT (ToolACE) initially suffers from distribution mismatch but later achieves substantial recovery after RL, indicating that RL can reactivate capabilities suppressed by divergent SFT data. When combined with ETS, the larger learning rate consistently improves all metrics, showing that error-driven supervision is highly sensitive to optimization scale. Overall, stabilizing multi-turn training may require larger learning rates than those commonly used in RL, which may explain instability in synchronous training under shared learning rates. 
\begin{table*}[!ht]
    \centering
    \small
    
    \renewcommand{\arraystretch}{1.2}
    \setlength{\tabcolsep}{6pt}
    \begin{tabular}{l ccc c c | ccc c c}
        \toprule
        \multirow{2}{*}{\textbf{Method}} 
        & \multicolumn{5}{c|}{\textbf{Format and Content OOD}} 
        & \multicolumn{5}{c}{\textbf{Content OOD}} \\
        \cmidrule(lr){2-6} \cmidrule(lr){7-11}
        & \textbf{Ad.} & \textbf{Sw.} & \textbf{sf.} & \textbf{Multi Turn} & \textbf{Avg}
        & \textbf{Ad.} & \textbf{Sw.} & \textbf{sf.} & \textbf{Multi Turn} & \textbf{Avg} \\
        \midrule

        Vanilla 
        & 38.0 & 20.0 & 46.0 & 0.0 & 26.00
        & 40.0 & 32.0 & 52.0 & 2.5 & 31.6 \\

        \rowcolor[HTML]{F1F8E9} $\text{SFT}_{\text{BFCL}}$ 
        & 0.0 & 0.0 & 0.0 & 0.0 & 0.00
        & \textbf{30.0} & 20.0 & 44.0 & 2.5 & 24.1 \\

        GRPO 
        & 32.0 & 24.0 & \textbf{43.0} & 0.0 & 24.75
        & 0.0 & 0.0 & 0.0 & 0.0 & 0.0 \\

        \rowcolor[HTML]{F1F8E9} $\text{SFT}_{\text{BFCL}}$ + RL 
        & 0.0 & 0.0 & 0.0 & 0.0 & 0.00
        & 8.0 & 8.0 & 10.0 & 0.0 & 6.5 \\

        OPS 
        & \textbf{36.0} & \textbf{24.0} & \textbf{43.0} & 0.0 & \textbf{25.75}
        & 20.0 & 16.0 & \textbf{50.0} & \textbf{6.7} & 23.2 \\

        HBG 
        & 28.0 & 20.0 & 42.0 & 0.0 & 22.50
        & 0.0 & 0.0 & 0.0 & 0.0 & 0.0 \\

        \rowcolor[HTML]{F1F8E9} ETS 
        & 10.0 & 0.0 & 25.0 & 0.0 & 8.75
        & 26.0 & 22.0 & 36.0 & 2.5 & 21.6 \\

        \rowcolor[HTML]{F1F8E9} PRS 
        & 16.0 & 8.0 & 24.0 & 0.0 & 12.00
        & 26.0 & \textbf{30.0} & 43.0 & 4.2 & \textbf{25.8} \\

        \bottomrule
    \end{tabular}
    \caption{Generalization results of Qwen2.5-1.5B-Instruct on ACEBench. Left: Format and Content OOD. Right: Content OOD. Best results are in bold, where \textbf{Ad.} denotes the Adjust subset, \textbf{Sw.} represents the Switch subset, \textbf{sf.} indicates single function. Green background means containing isolated SFT stage.}
    \label{tab:merged_ood_results}
\end{table*}

\subsection{The Analysis of Generalization}
We evaluate model generalization under out-of-distribution (OOD) settings using the original ACEBench framework, which features unseen tool types and prompt structures. Many prior studies prioritize scenario generalization without format generalization. We report both generalization axes to analyze the role of supervisory signals. 

\textbf{Defining ID vs. OOD in our setup.} We distinguish content OOD from format OOD. Content OOD evaluates models on tool types and scenarios absent from the in-distribution (ID) training data. Format OOD is instantiated by ACEBench’s fixed invocation template, whereas ID evaluation (including training) follows Qwen’s native format. Therefore, we observe a general performance drop in OOD scenarios, caused by both factors. Experimental results yield several critical insights:

\textbf{Format- and content-sensitive overfitting during SFT.} In Table~\ref{tab:merged_ood_results}, SFT methods that excel in-distribution suffer sharp degradation under Format and Content OOD in simple single-step settings but not on Content OOD. This supports our claim that standard SFT induces severe format overfitting, tethering the model to training-time syntax. 

\textbf{RL ``collapse'' as a format artefact}. Methods that appear unstable during training (e.g., GRPO-only, OPS, HBG) exhibit more stable behavior under Format and Content OOD tests, while Content OOD in same tool format experiments show decline. This confirms that the observed instability is format-specific (token-level probability shifts) rather than a global loss of reasoning or linguistic competence, as hypothesized in Section~\ref{sec:analysis}. The Multi Turn results further indicate limited generalization during training. 

\textbf{Process-Level Regularization}. While interleaved training can degrade single-turn performance, the integration of Process Reflection Supervision (PRS) effectively mitigates this decline. In addition, these results indicate that isolated SFT stages are more prone to inducing format overfitting: while they can substantially improve in-distribution performance, they often harm generalization unless complemented with process-level supervision. Unlike isolated SFT which prioritizes output-format optimization, PRS encourages the model to internalize the underlying logical scaffolding of tool use. This suggests that incorporating reasoning-heavy, process-oriented data during the SFT phase acts as a crucial logical regularizer, enhancing the model's resilience to distribution shifts in real-world agentic environments.

\section{Conclusion}
This work examines reinforcement learning for multi-turn tool-use, identifying two key failure modes: training instability and limited gains. Naive RL can over-amplify control tokens, disrupting structured tool use and causing collapse or plateaus. We systematically study supervisory signal integration, finding interleaved training more stable and effective than synchronous. We also analyze learning rate, SFT data distribution, and generalization, offering a principled framework for stable, effective agentic RL in tool-use settings.

\section*{Limitations}
This work indicates and analyzes failure modes in multi-turn tool-use tasks, highlighting the critical role of specific tokens and investigating how different intervention signals help mitigate these issues. However, the training data used in this study is relatively limited due to the limited open-source verifiable tool invoking environment, and the impact of data scale on the results is not explored.

\section*{Ethics Statement}
Our work does not introduce ethical concerns. This paper utilized AI assistance for language polishing of the manuscript, including vocabulary correction and spell checking. 

\bibliography{custom}

\appendix
\newpage
\section{Analysis}
\label{appendix:analysis}
\begin{figure*}[t]
    \centering
    \includegraphics[width=1\textwidth]{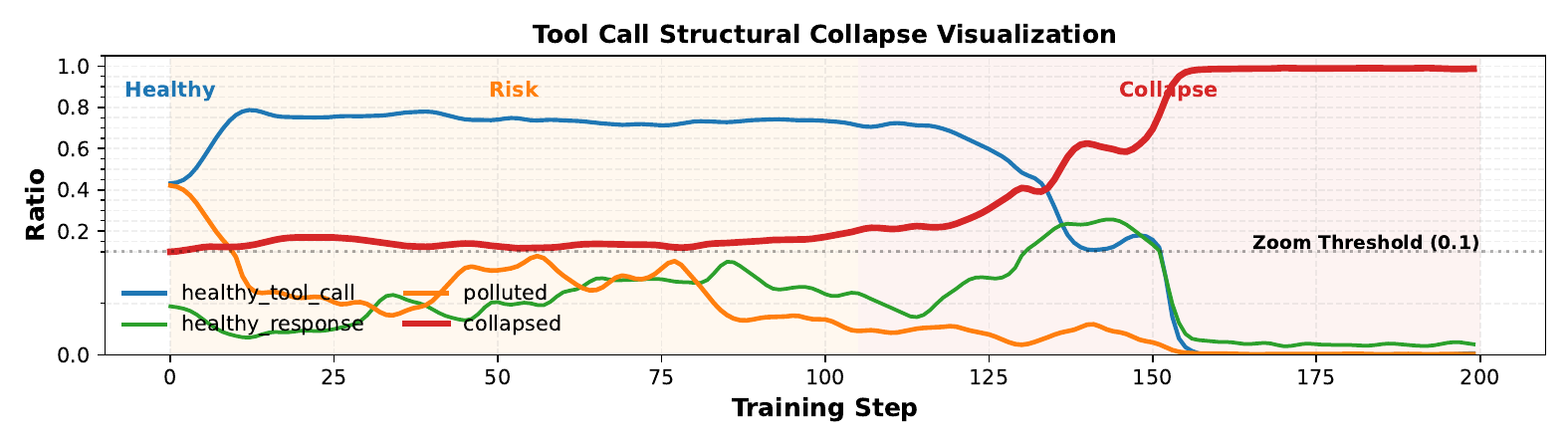}
    \caption{The changes of token appearing frequency in the SFT then RL training process of model Qwen3-1.7B.}
    \label{fig:token_probility_3}
\end{figure*}

\begin{figure*}[t]
    \centering
    \includegraphics[width=1\textwidth]{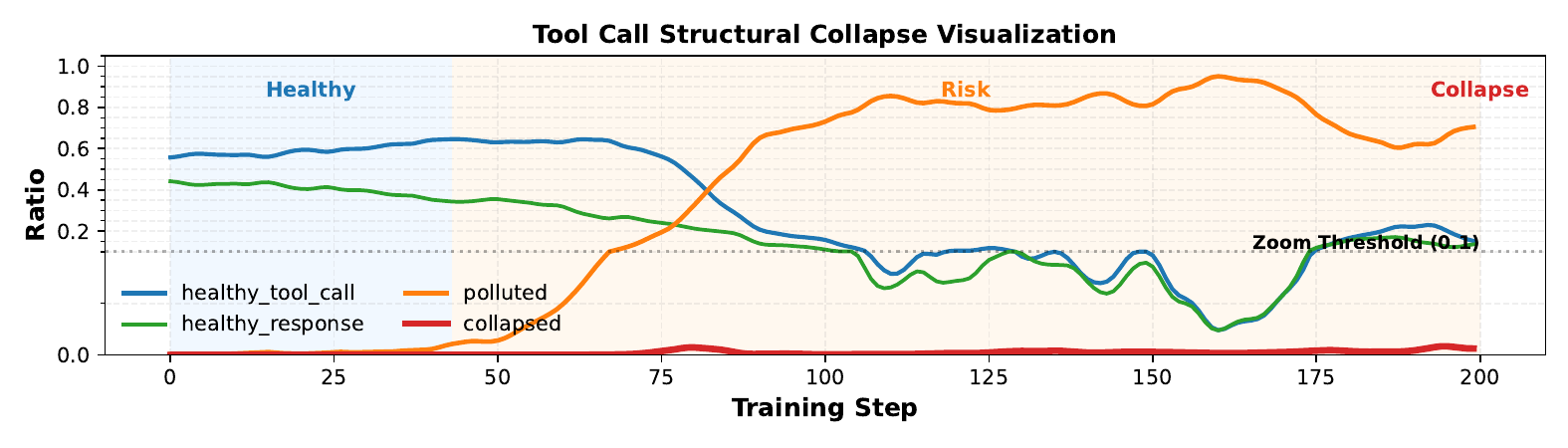}
    \caption{The changes of token appearing frequency in the RL training process of model Qwen3-1.7B.}
    \label{fig:token_probility_3_rl}
\end{figure*}
\begin{table*}[!ht]
    \centering
    \small
    \setlength{\tabcolsep}{10pt}
    
    \resizebox{\textwidth}{!}{
    \renewcommand{\arraystretch}{1.2}
    \begin{tabular}{l c}
        \toprule
        \textbf{Type} & \textbf{Example} \\
        \midrule
        Healthy Tool Call 
        & \texttt{<tool\_call>\{``name'': ``search'', ``arguments'': \{...\}\}</tool\_call><|im\_end|>} \\

        Healthy Response 
        & \texttt{The answer can be derived by first querying the database...<|im\_end|>} \\

        Polluted Output 
        & \texttt{The result is ...<tool\_call><|im\_end|> } \\

        Collapsed Output 
        & \texttt{<tool\_call><|im\_end|>} \\
        \bottomrule
    
    \end{tabular}}
    \caption{Examples of different output types observed during multi-turn tool-use training.}
    \label{tab:instance}
\end{table*}
The detailed classification of model output structures:
\begin{itemize}
    \item Healthy Tool Call: Model output contains a valid tool invocation: both opening \texttt{<tool\_call>} and closing \texttt{</tool\_call>} tags are present and correctly matched. The internal action is valid, the structure conforms to the expected JSON schema, and no stray \texttt{<|im\_end|>} tokens or partial tags appear within the invocation. 
    \item Healthy Response: Model output does \textbf{not} include any tool-call tags (\texttt{<tool\_call>}, \texttt{</tool\_call>}) but ends with a valid end-of-turn token \texttt{<|im\_end|>}. The content is a coherent natural language response without extraneous or misused tool tags. \\
    \item Text Polluted: Outputs with partially misused tool-call tags, such as missing closing tags, adding \texttt{<think>}, embedded \texttt{<|im\_end|>} inside a tool invocation, or extra tags appended to normal text. Indicates structural corruption where the model mixes dialogue text with tool invocation tags incorrectly, potentially causing protocol violations or parsing errors. \\
    \item Collapsed: Outputs consisting solely of the end-of-turn token \texttt{<|im\_end|>}, or cases where all tool invocations collapse to just \texttt{<|im\_end|>} without any semantic content. Reflects severe RL failure, where the model short-circuits the interaction to maximize reward or avoid penalties, producing no meaningful action or response. \\
\end{itemize}
 The more analysis results are in Figure~\ref{fig:token_probility_3} and Figure~\ref{fig:token_probility_3_rl}.

\section{Training Details}
\label{appendix:training_details}
During RL training for Qwen3, we prepend \texttt{<think>\textbackslash n\textbackslash n</think>\textbackslash n\textbackslash n} before each action to disable the thinking mode. During gradient updates, this prefix is removed, and only the generated content is used for optimization. 
\begin{table}[!ht]
\centering

\renewcommand{\arraystretch}{1.2}
\setlength{\tabcolsep}{8pt}
\begin{tabular}{l l}
\toprule
\textbf{Parameter} & \textbf{SFT} \\
\midrule
Epochs & 3 \\
Train batch size & 128 \\
Learning rate & 1e-5 \\
Max sequence length & 8192 \\
\bottomrule
\end{tabular}
\caption{Supervised Fine-Tuning (SFT) configuration.}
\label{tab:sft_config}
\end{table}

\begin{table}[!ht]
\centering
\renewcommand{\arraystretch}{1.2}

\setlength{\tabcolsep}{8pt}
\begin{tabular}{l l}
\toprule
\textbf{Parameter} & \textbf{RL} \\
\midrule
Total training steps & 300 \\
Batch size & 60 \\
PPO mini-batch size & 20 \\
Learning rate & 1e-6 \\
Max turns per trajectory & 30 \\
KL loss coefficient & 0.0001 \\
\bottomrule
\end{tabular}
\caption{Reinforcement Learning (RL) configuration.}
\label{tab:rl_config_core}
\end{table}
The SFT and RL parameter configurations are summarized in the Table~\ref{tab:sft_config} and Table~\ref{tab:rl_config_core}. The experimental setups for the different supervisory interventions are as follows: for off-policy supervision, 7 trajectories are sampled from the model and 1 trajectory uses the ground-truth label; for hint-based guidance, 6 trajectories are generated normally and 2 are conditioned on hints. For interleaved training, the RL rounds for Qwen2.5-1.5B-Instruct are 50, 100, and 150, while for Qwen3-1.7B, they are 50, 50, and 100.

We follow the BFCL reward design in all the reinforcement learning with two evaluation types: \textbf{state-based} and \textbf{response-based} evaluation. The reward is 1 if both state-based evaluation and response-based evaluation is 1, and it will be 0 if not. State-based evaluation compares the backend state after all function calls with the ground-truth final state, focusing on tasks that modify internal states. Response-based evaluation compares the model's execution path with the minimal viable ground-truth function-call path, enabling evaluation of read-only tasks such as querying stock or weather information.

In RPS, we use gpt-5-mini as the auxiliary analyzer to generate textual reflections. The prompt and output example are in Appendix~\ref{sec:appendix_prompt_and_example}.

\section{The Analysis of Qwen3 Training}

\label{appendix:influence_qwen3_prompt}
\begin{figure*}[t]
    \centering
    \includegraphics[width=1\textwidth]{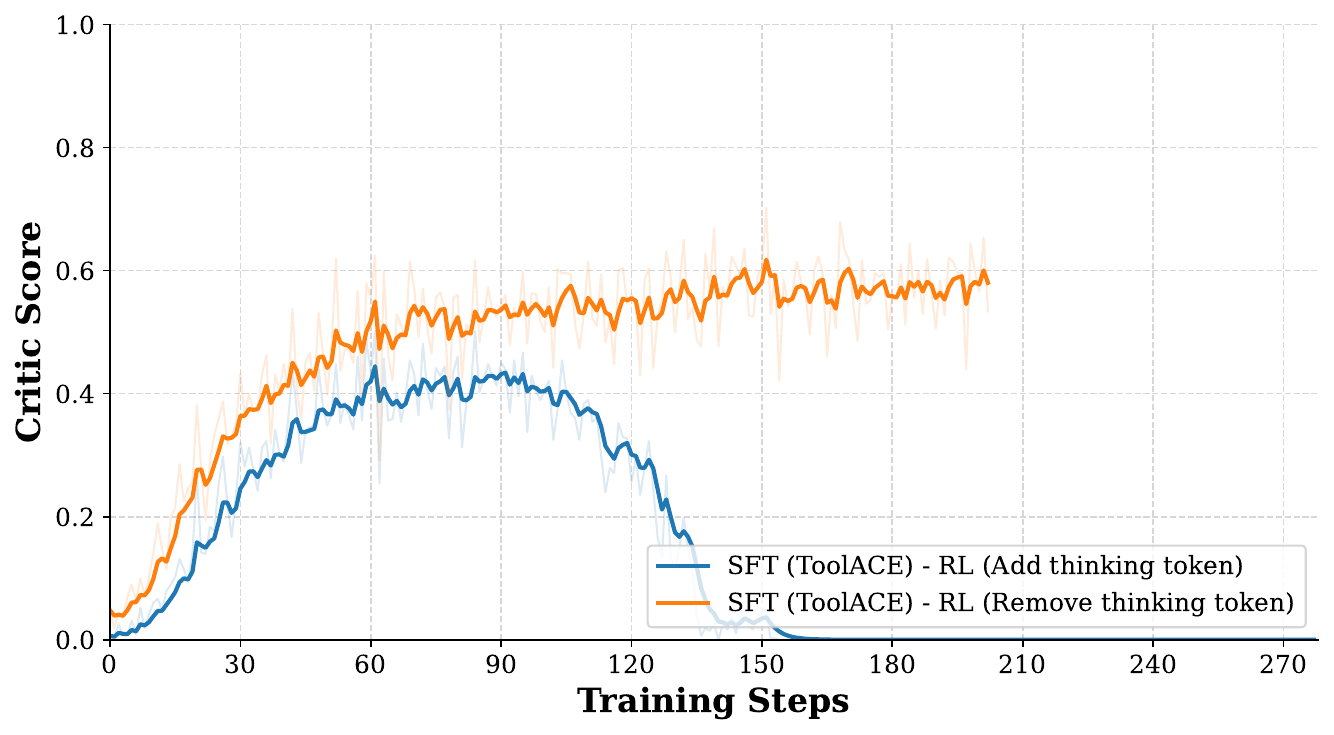}
    \caption{The training reward with or without thinking token}
    \label{fig:reward_qwen3}
\end{figure*}

We believe that the performance collapse observed in Qwen3 after the SFT-then-RL pipeline is primarily caused by inconsistencies in prompt formatting. Specifically, Qwen3 is designed to operate in a thinking mode, where the model is expected to generate an explicit reasoning segment. When reasoning output is disabled, the prompt must explicitly include a placeholder such as \texttt{<think>\textbackslash n\textbackslash n</think>}.
However, during the SFT stage, the training data contains only direct tool-call outputs without this reasoning wrapper. If, during the subsequent RL stage, the prompt is modified to include the thinking-related tokens while the supervised data does not reflect this structure, a mismatch is introduced between the SFT distribution and the RL sampling format. This inconsistency can destabilize training, leading to structural drift and eventual performance collapse.
The results in Figure~\ref{fig:reward_qwen3} demonstrate this point.

\section{Detailed Evaluation}
\label{appendix:detailed_evaluation}

\begin{figure*}[ht]
    \centering
    \begin{subfigure}[t]{0.24\linewidth}
        \centering
        \includegraphics[width=\linewidth]{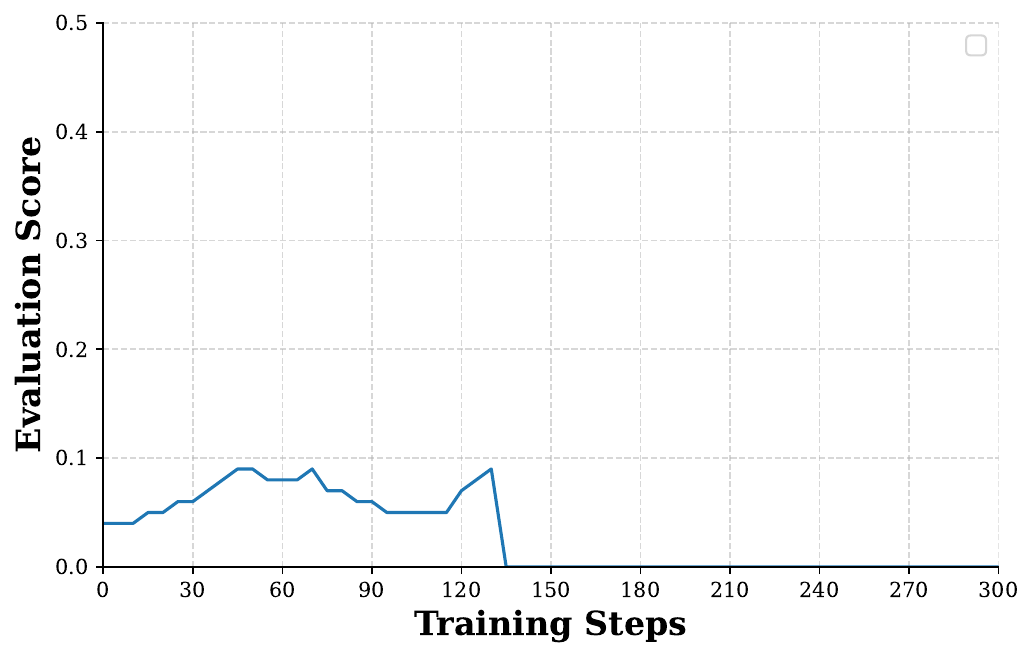}
        \caption{RL}
    \end{subfigure}
    \hfill
    \begin{subfigure}[t]{0.24\linewidth}
        \centering
        \includegraphics[width=\linewidth]{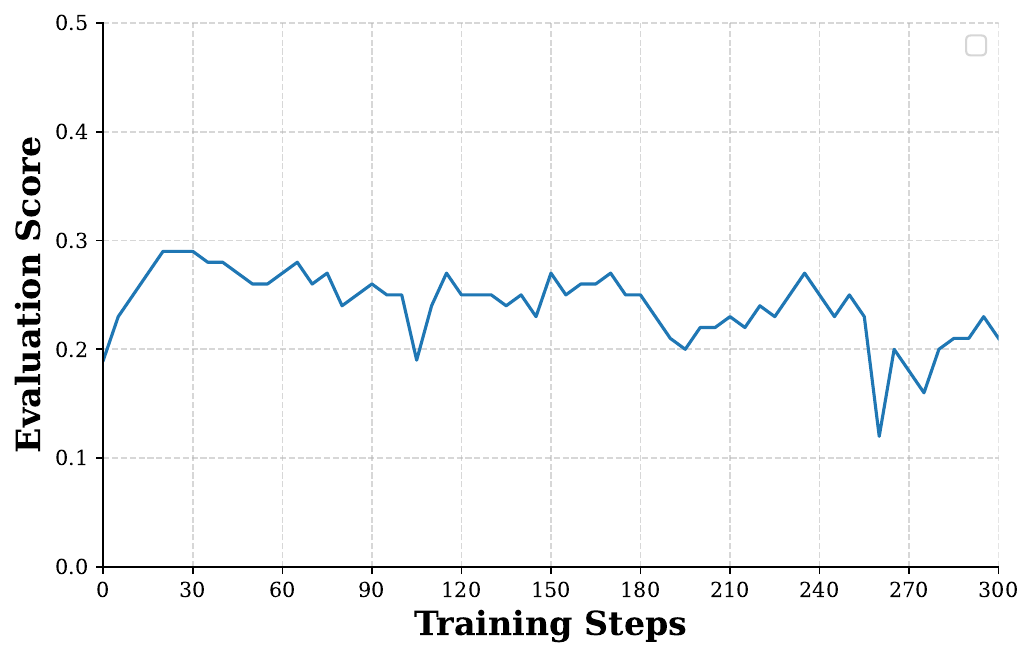}
        \caption{SFT (BFCL) + RL}
    \end{subfigure}
    \hfill
    \begin{subfigure}[t]{0.24\linewidth}
        \centering
        \includegraphics[width=\linewidth]{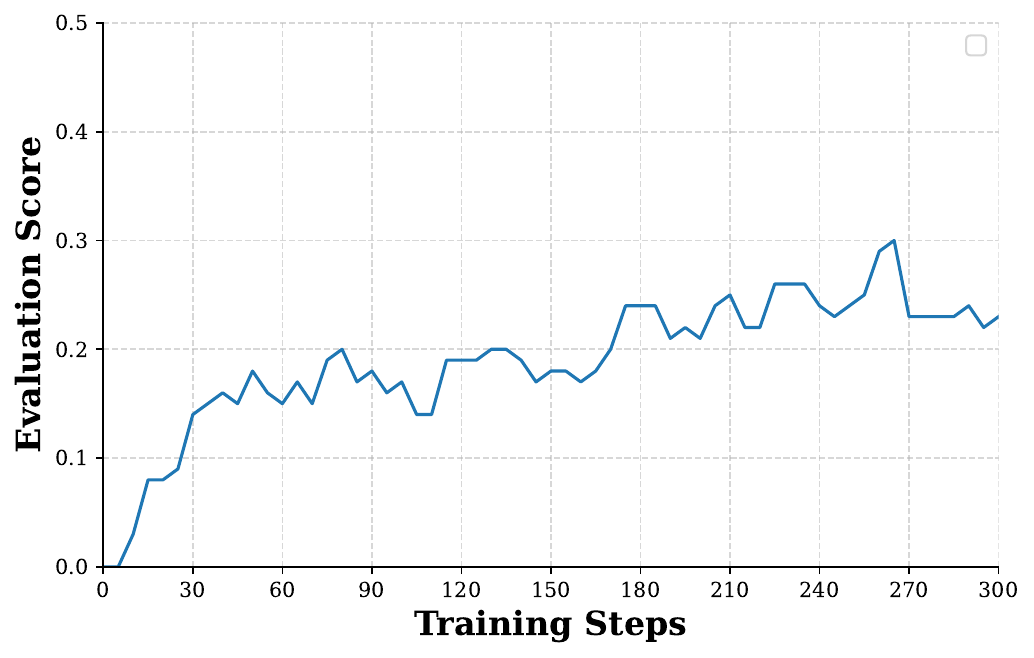}
        \caption{SFT (ToolACE) + RL}
    \end{subfigure}
    \hfill
    \begin{subfigure}[t]{0.24\linewidth}
        \centering
        \includegraphics[width=\linewidth]{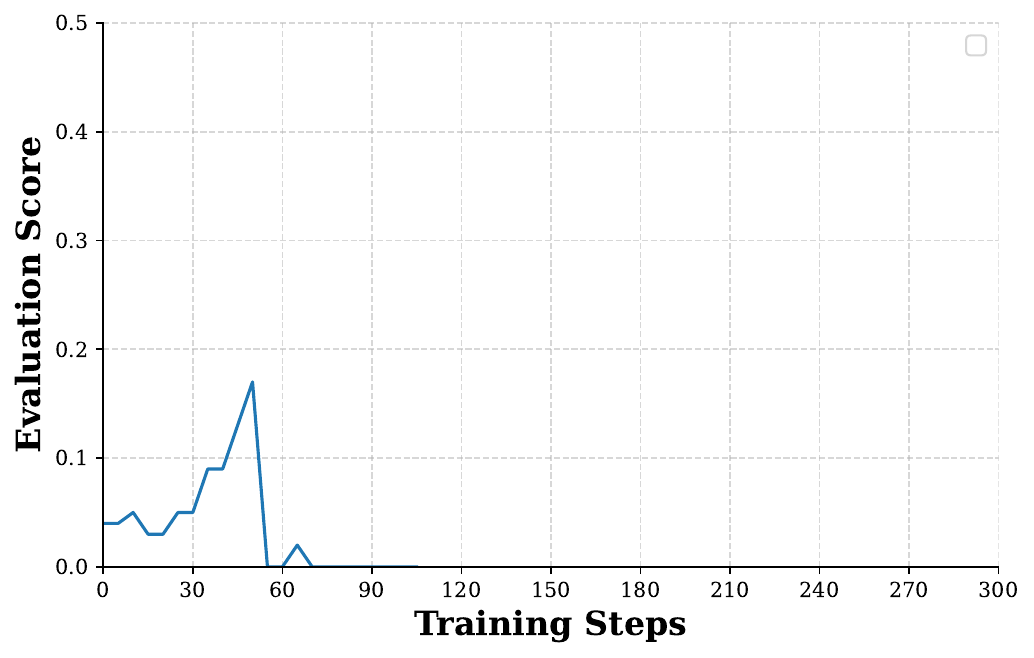}
        \caption{OPS}
    \end{subfigure}

    \begin{subfigure}[t]{0.24\linewidth}
        \centering
        \includegraphics[width=\linewidth]{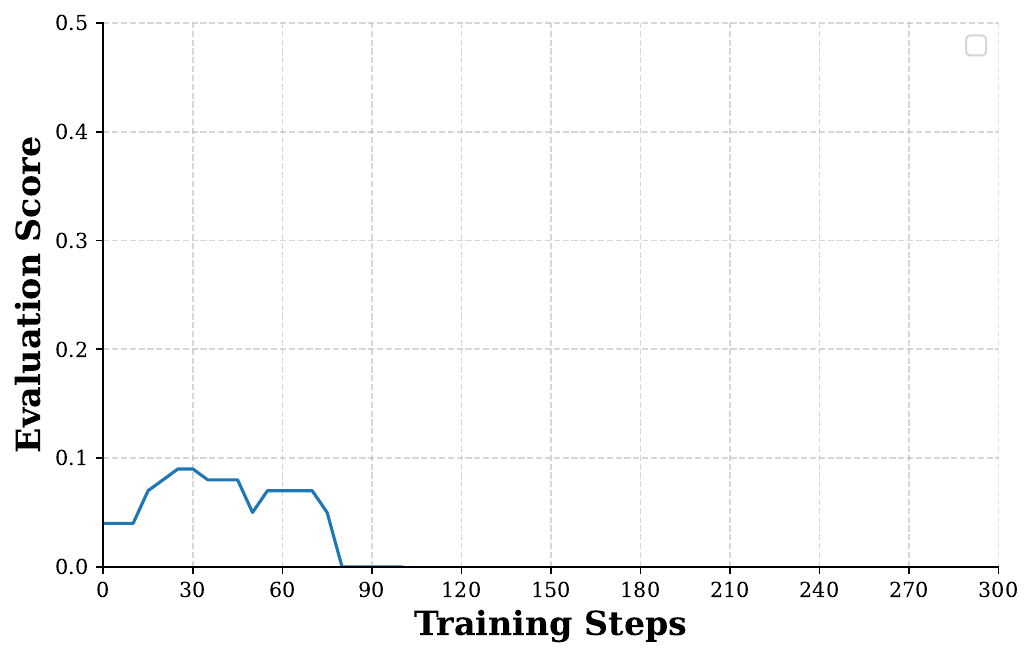}
        \caption{HBG}
    \end{subfigure}
    \hfill
    \begin{subfigure}[t]{0.24\linewidth}
        \centering
        \includegraphics[width=\linewidth]{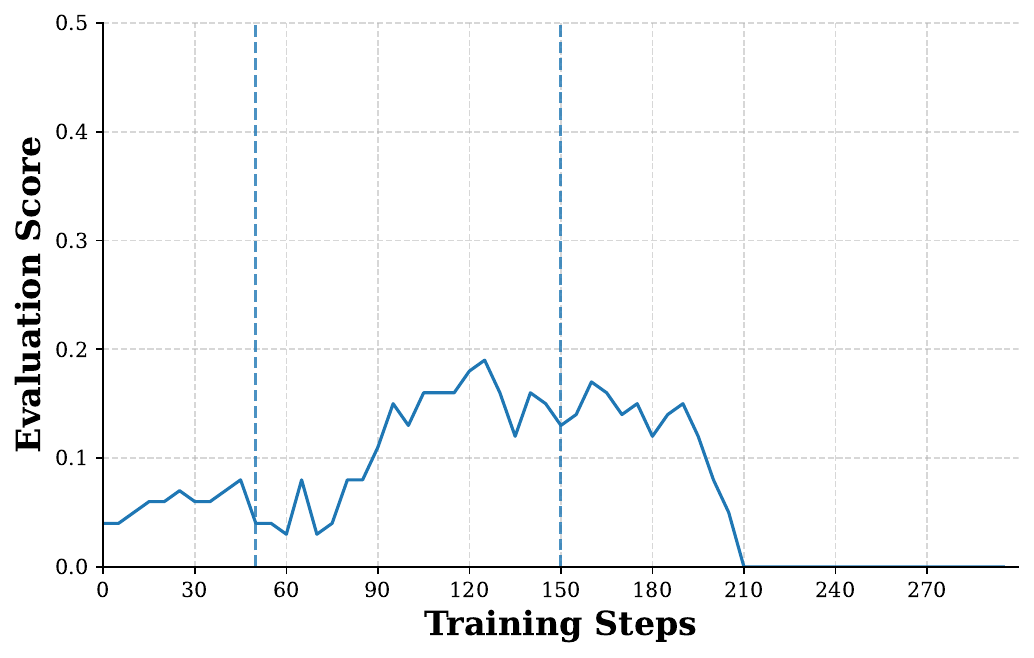}
        \caption{ETS (SFT $1\mathrm{e}{-6}$)}
    \end{subfigure}
    \hfill
    \begin{subfigure}[t]{0.24\linewidth}
        \centering
        \includegraphics[width=\linewidth]{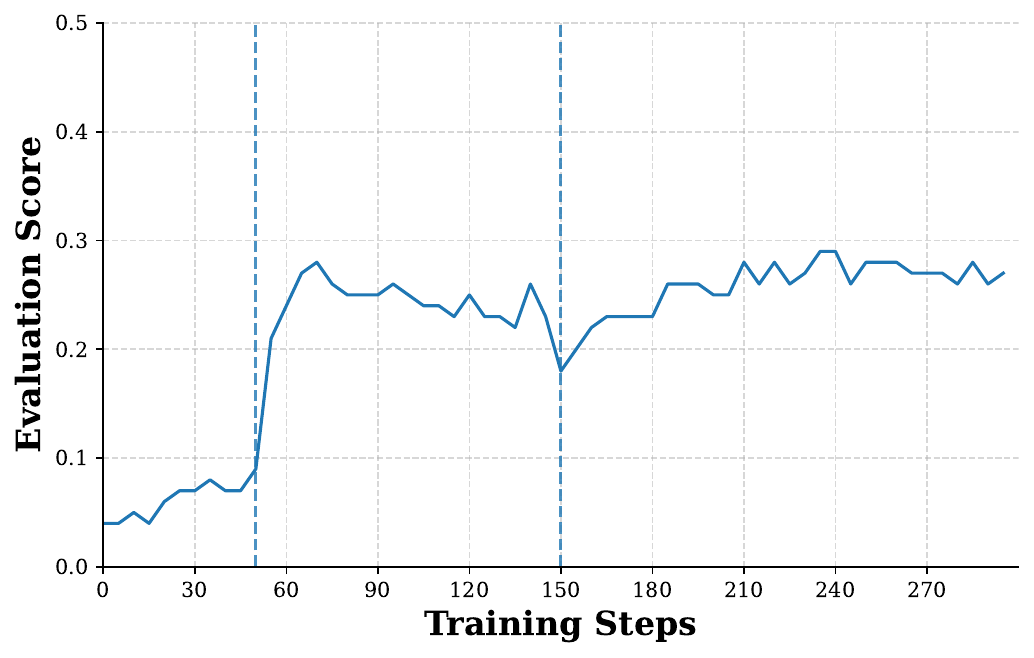}
        \caption{ETS (SFT $1\mathrm{e}{-5}$)}
    \end{subfigure}
    \hfill
    \begin{subfigure}[t]{0.24\linewidth}
        \centering
        \includegraphics[width=\linewidth]{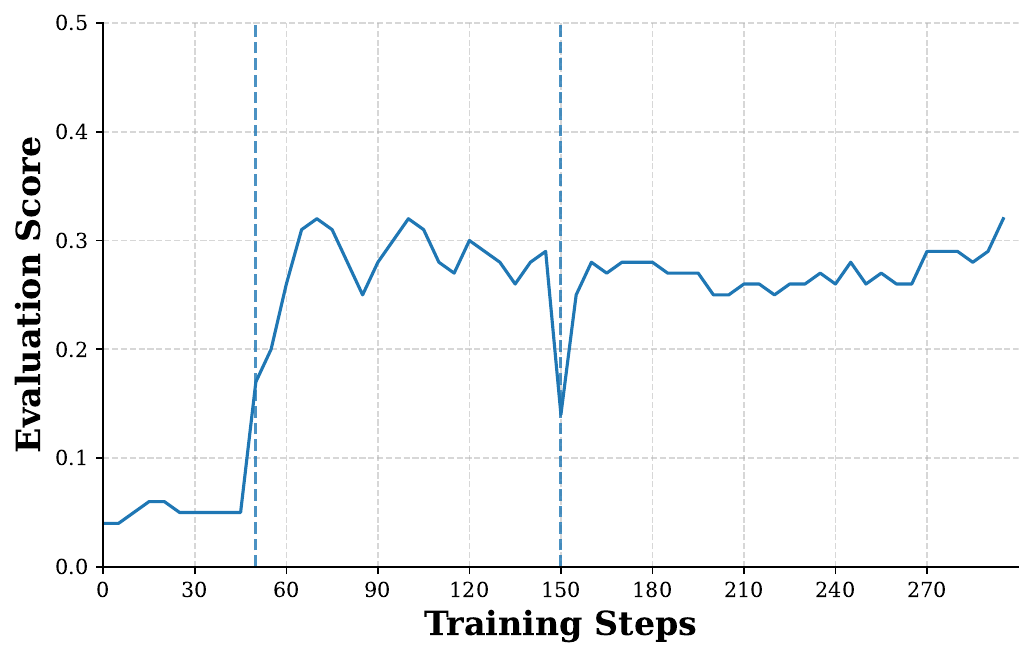}
        \caption{PRS}
    \end{subfigure}

    \caption{Evaluation results of scenario \textit{Base} under different supervisory signals of Qwen2.5-1.5B-Instruct.}
    \label{fig:evaluation_results_base}
\end{figure*}
\begin{figure*}[ht]
    \centering
    \begin{subfigure}[t]{0.24\linewidth}
        \centering
        \includegraphics[width=\linewidth]{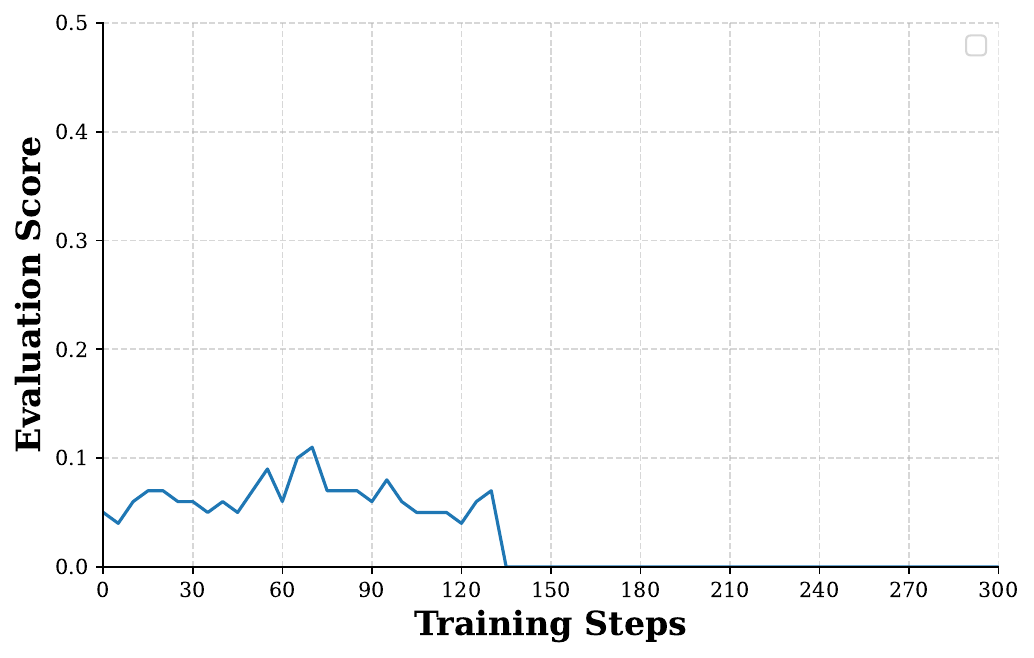}
        \caption{RL}
    \end{subfigure}
    \hfill
    \begin{subfigure}[t]{0.24\linewidth}
        \centering
        \includegraphics[width=\linewidth]{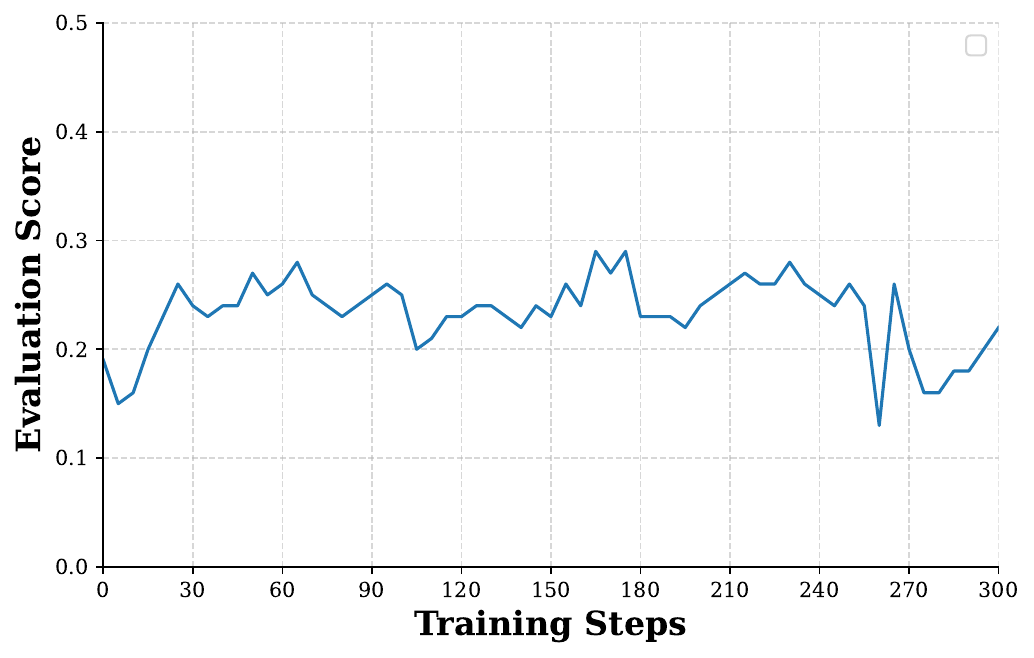}
        \caption{SFT (BFCL) + RL}
    \end{subfigure}
    \hfill
    \begin{subfigure}[t]{0.24\linewidth}
        \centering
        \includegraphics[width=\linewidth]{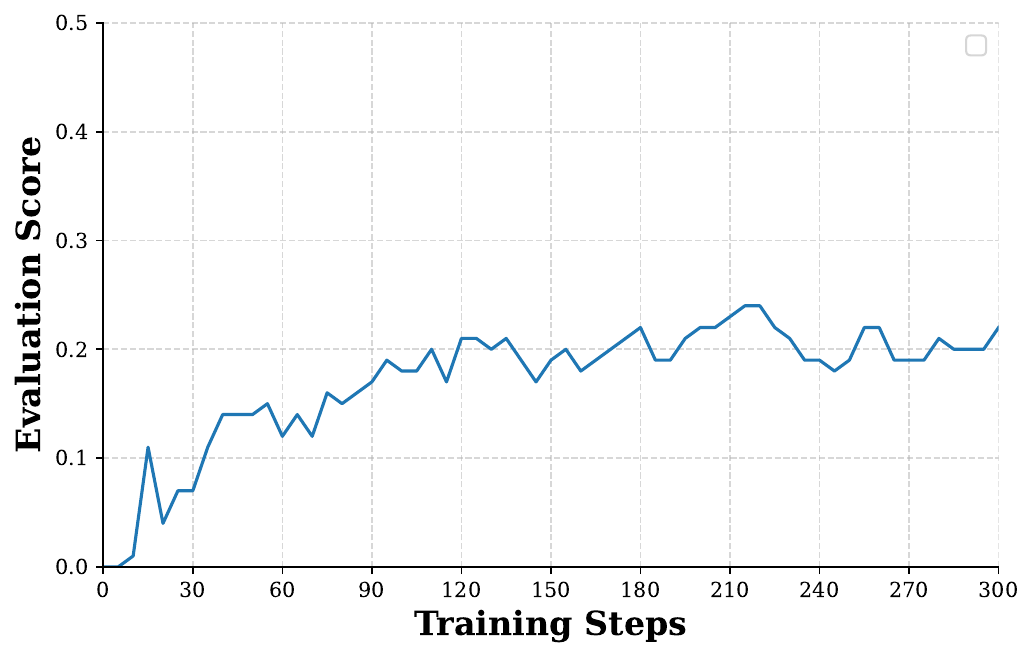}
        \caption{SFT (ToolACE) + RL}
    \end{subfigure}
    \hfill
    \begin{subfigure}[t]{0.24\linewidth}
        \centering
        \includegraphics[width=\linewidth]{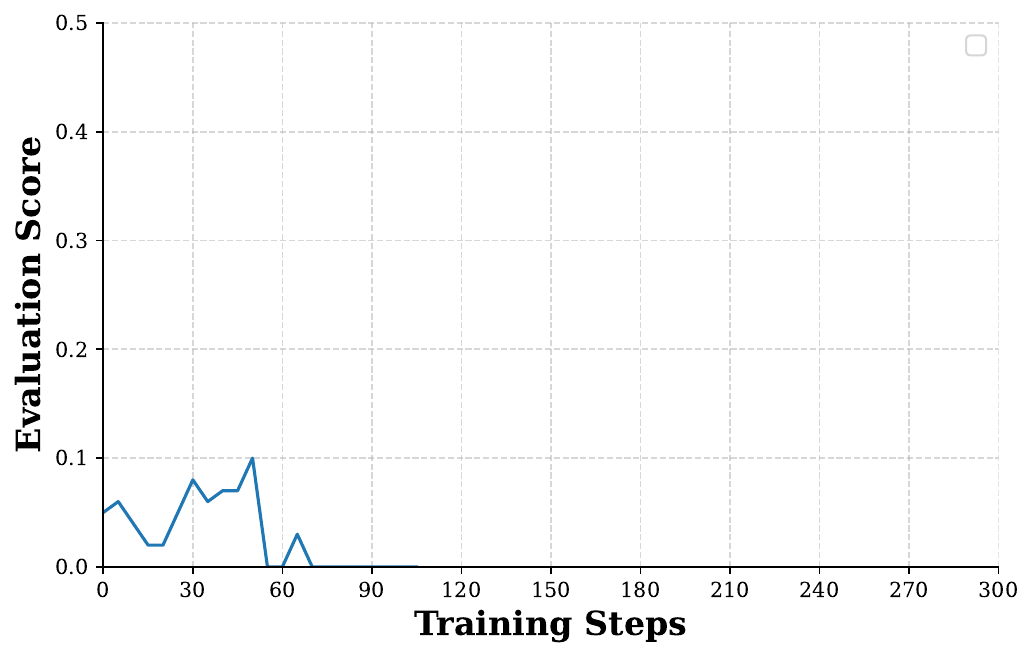}
        \caption{OPS}
    \end{subfigure}

    \begin{subfigure}[t]{0.24\linewidth}
        \centering
        \includegraphics[width=\linewidth]{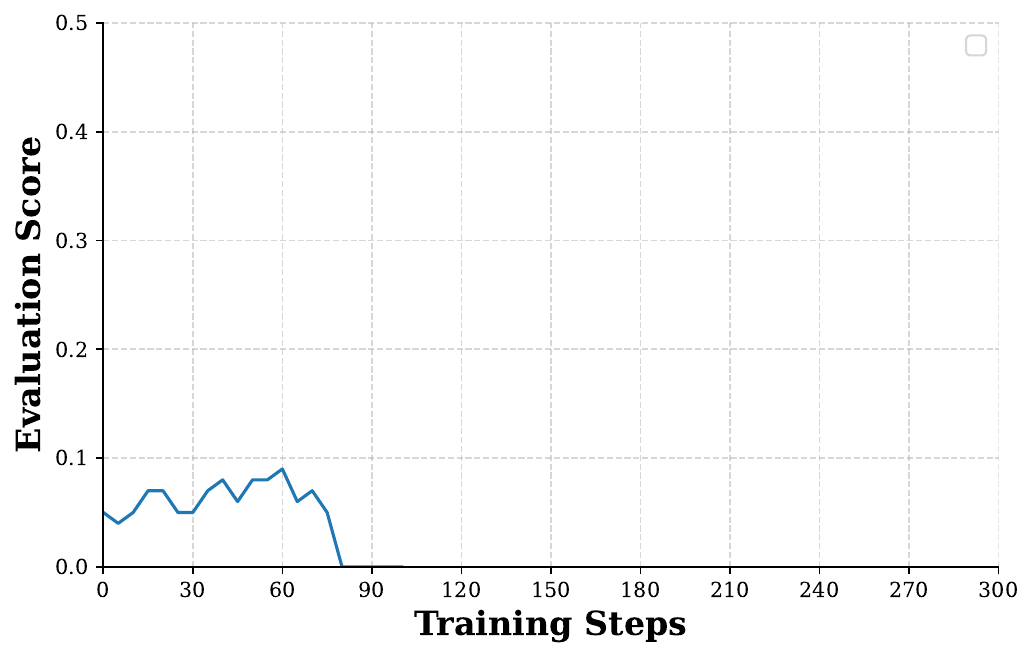}
        \caption{HBG}
    \end{subfigure}
    \hfill
    \begin{subfigure}[t]{0.24\linewidth}
        \centering
        \includegraphics[width=\linewidth]{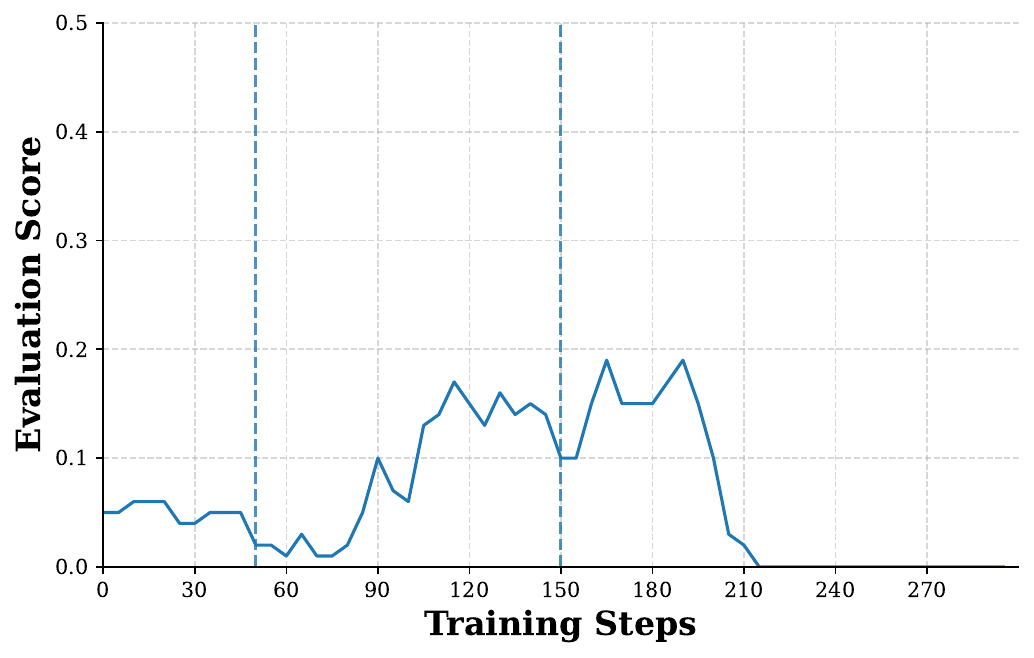}
        \caption{ETS (SFT $1\mathrm{e}{-6}$)}
    \end{subfigure}
    \hfill
    \begin{subfigure}[t]{0.24\linewidth}
        \centering
        \includegraphics[width=\linewidth]{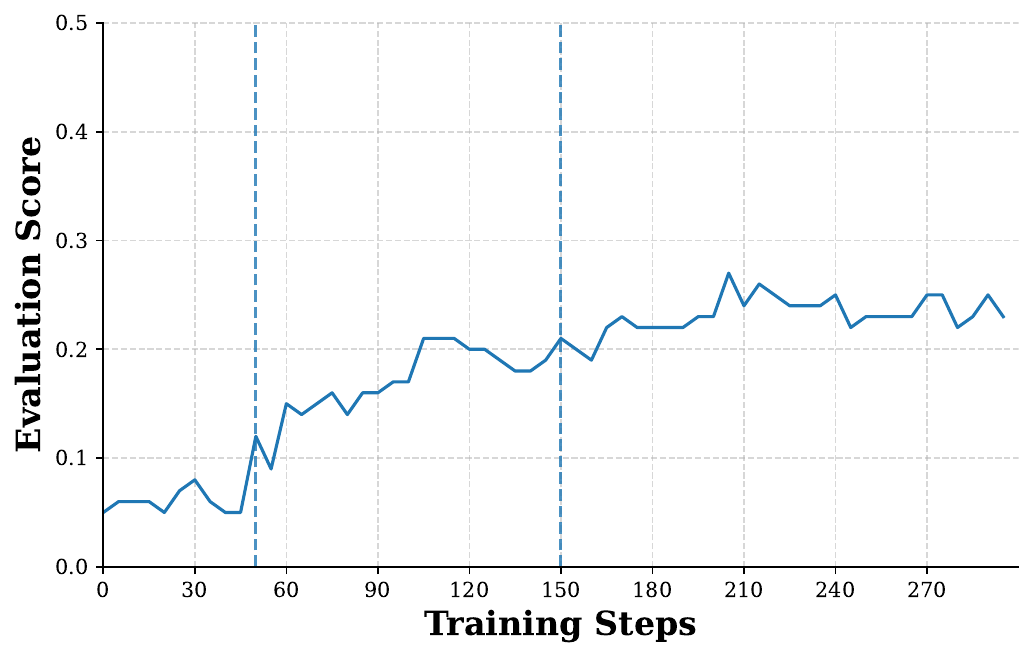}
        \caption{ETS (SFT $1\mathrm{e}{-5}$)}
    \end{subfigure}
    \hfill
    \begin{subfigure}[t]{0.24\linewidth}
        \centering
        \includegraphics[width=\linewidth]{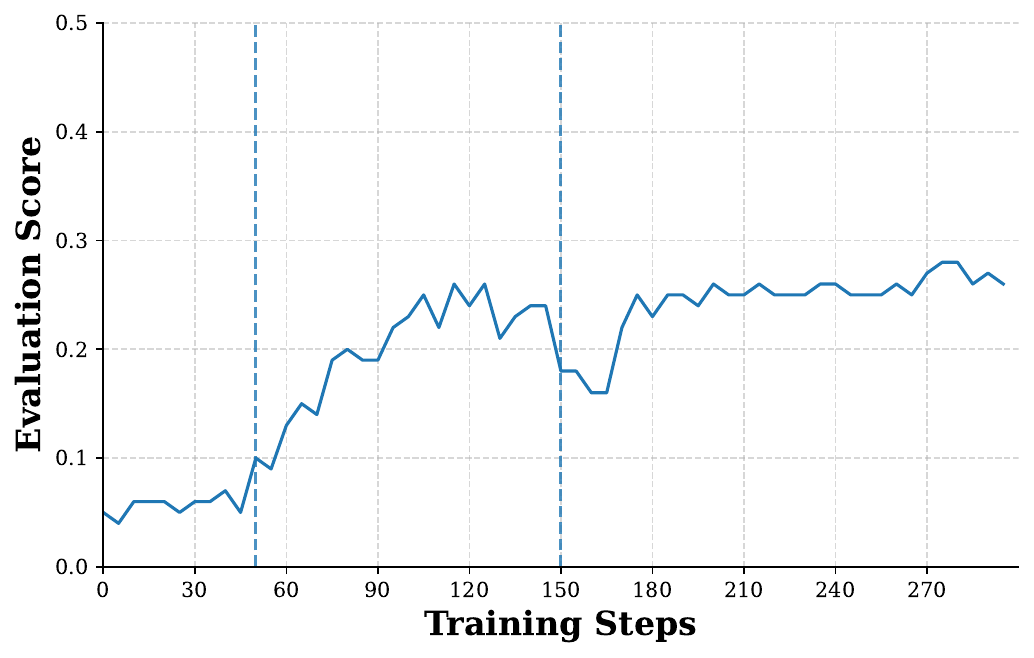}
        \caption{PRS}
    \end{subfigure}

    \caption{Evaluation results of scenario \textit{Miss Func} under different supervisory signals of Qwen2.5-1.5B-Instruct.}
    \label{fig:evaluation_results_miss_func}
\end{figure*}
\begin{figure*}[ht]
    \centering
    \begin{subfigure}[t]{0.24\linewidth}
        \centering
        \includegraphics[width=\linewidth]{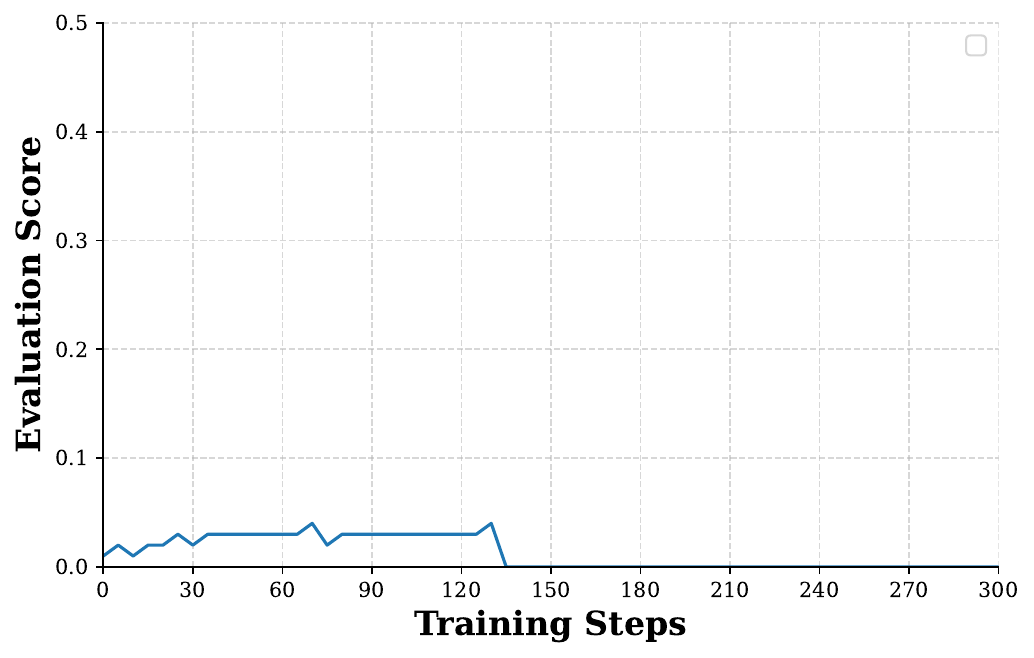}
        \caption{RL}
    \end{subfigure}
    \hfill
    \begin{subfigure}[t]{0.24\linewidth}
        \centering
        \includegraphics[width=\linewidth]{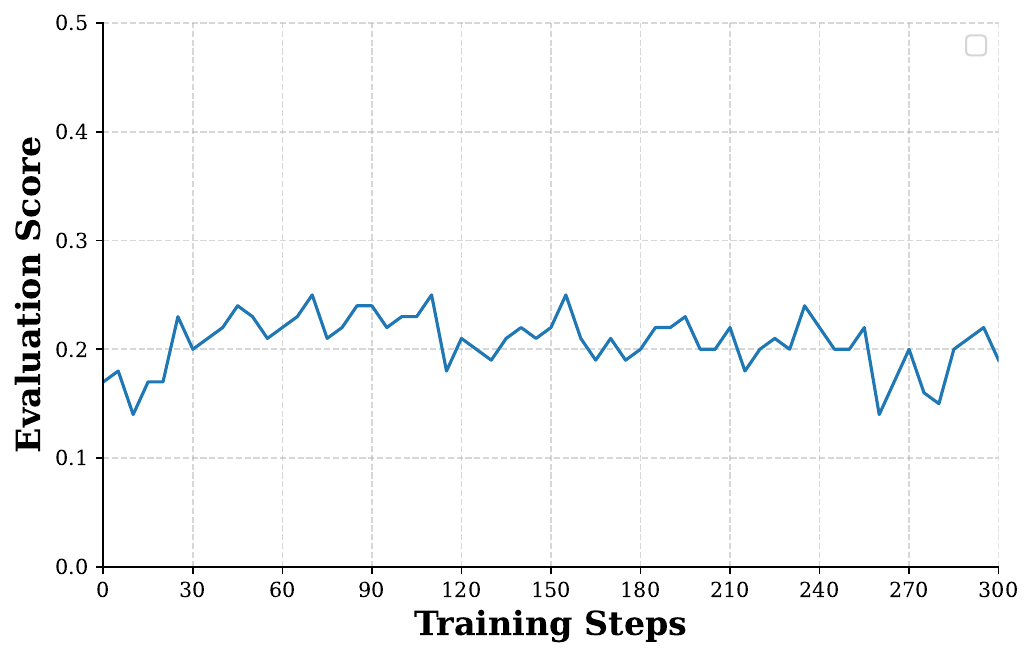}
        \caption{SFT (BFCL) + RL}
    \end{subfigure}
    \hfill
    \begin{subfigure}[t]{0.24\linewidth}
        \centering
        \includegraphics[width=\linewidth]{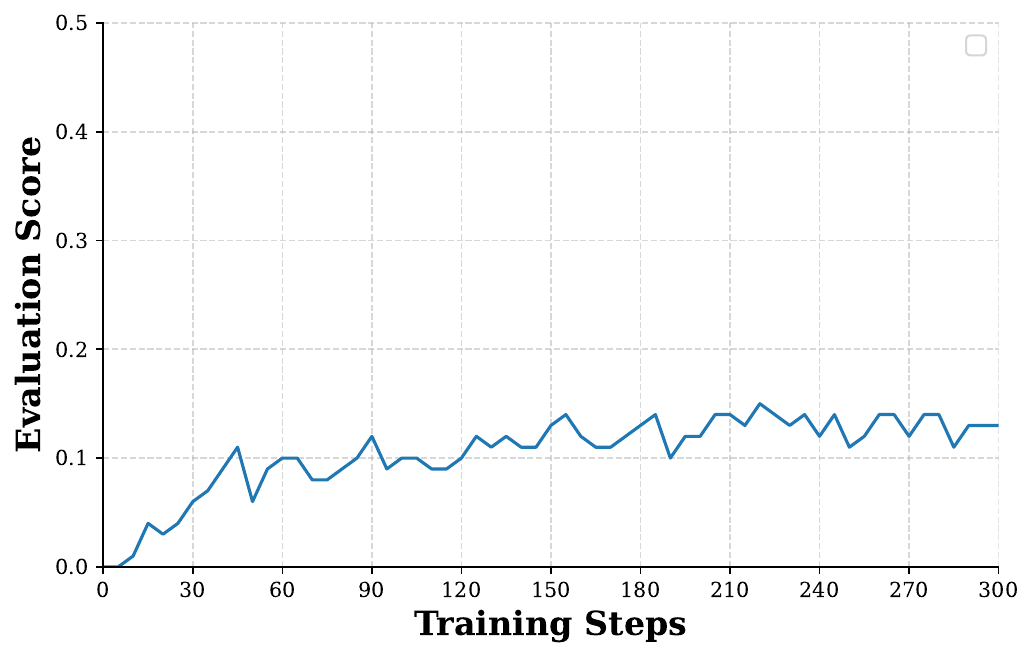}
        \caption{SFT (ToolACE) + RL}
    \end{subfigure}
    \hfill
    \begin{subfigure}[t]{0.24\linewidth}
        \centering
        \includegraphics[width=\linewidth]{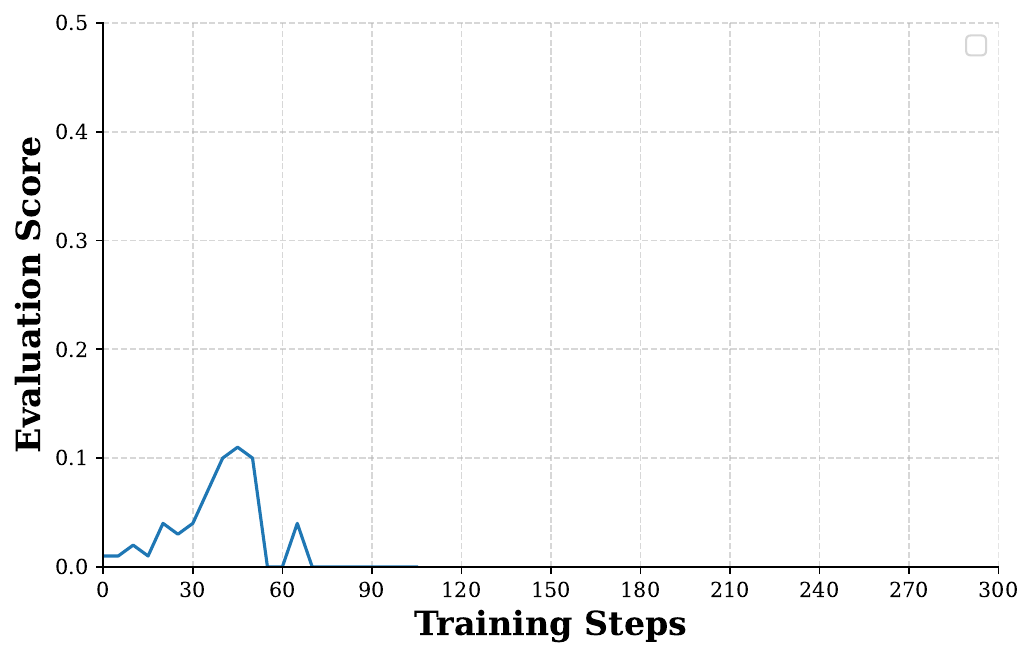}
        \caption{OPS}
    \end{subfigure}

    \begin{subfigure}[t]{0.24\linewidth}
        \centering
        \includegraphics[width=\linewidth]{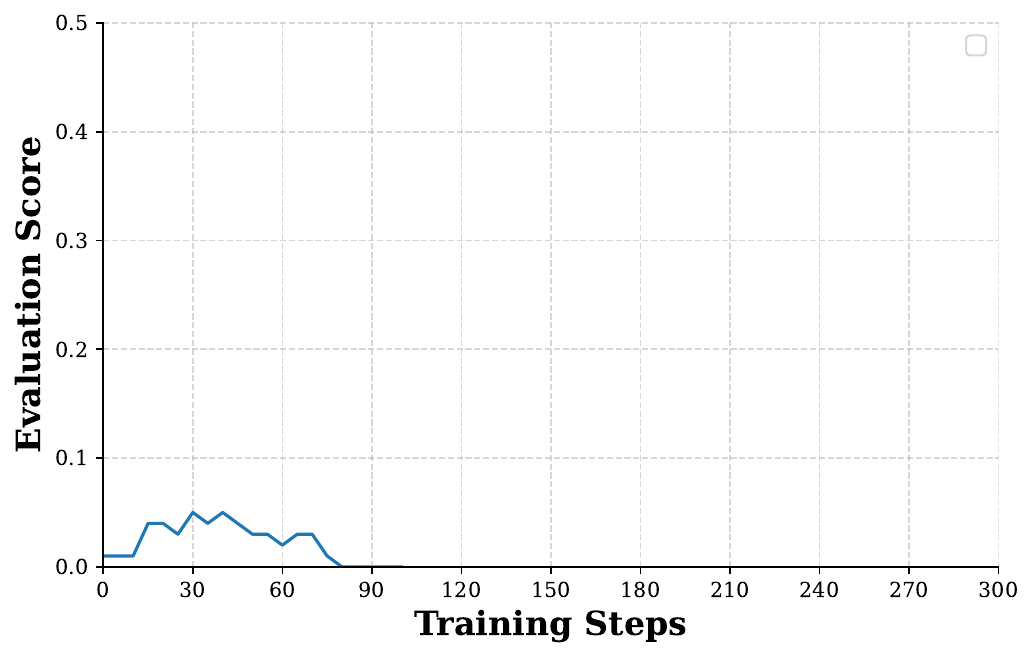}
        \caption{HBG}
    \end{subfigure}
    \hfill
    \begin{subfigure}[t]{0.24\linewidth}
        \centering
        \includegraphics[width=\linewidth]{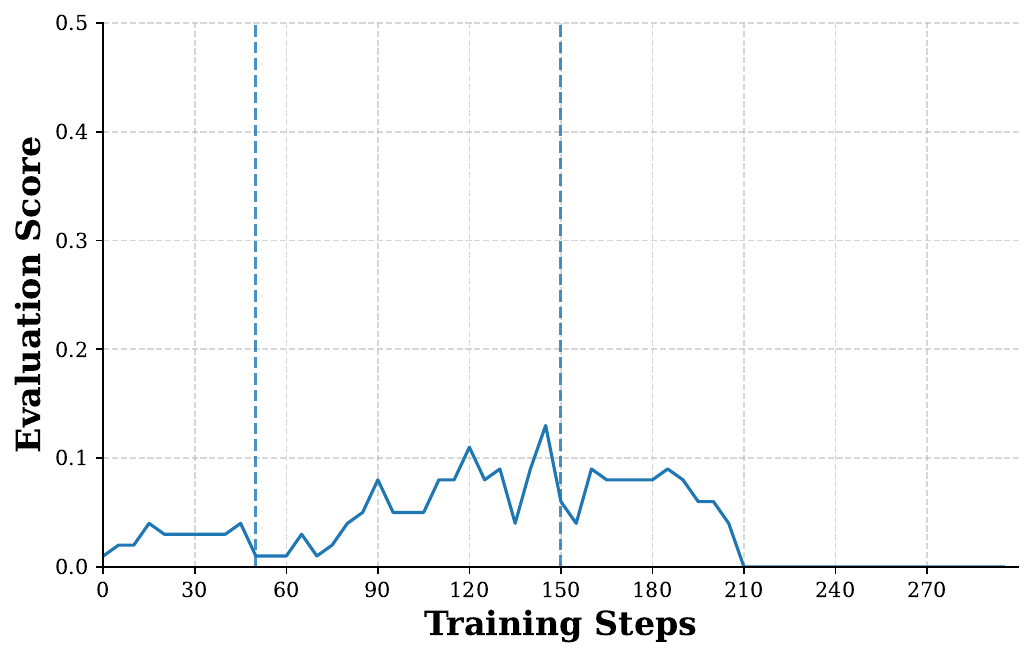}
        \caption{ETS (SFT $1\mathrm{e}{-6}$)}
    \end{subfigure}
    \hfill
    \begin{subfigure}[t]{0.24\linewidth}
        \centering
        \includegraphics[width=\linewidth]{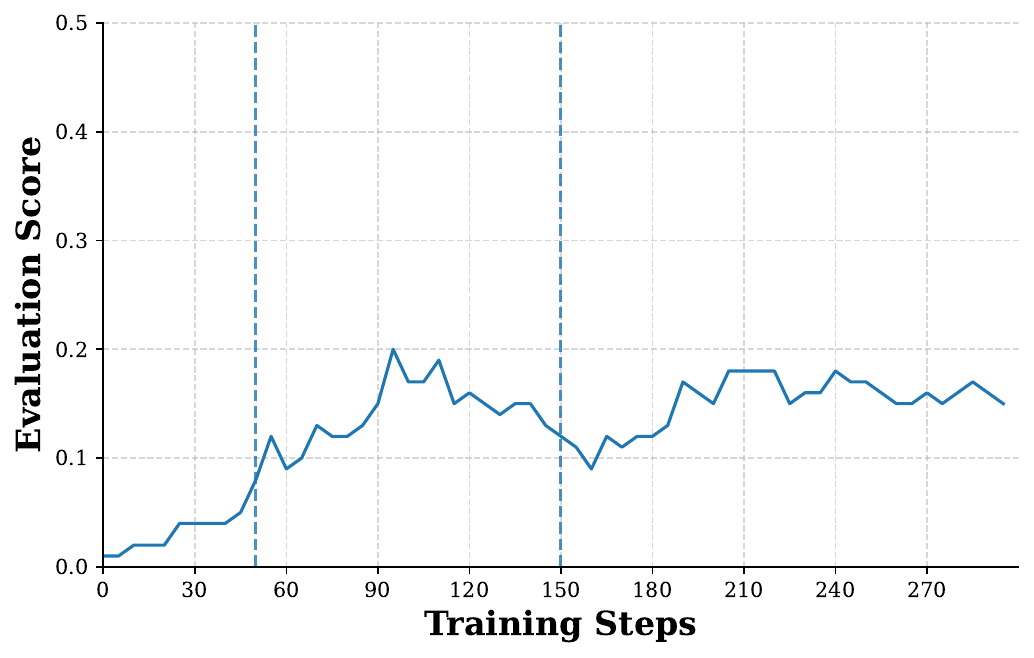}
        \caption{ETS (SFT $1\mathrm{e}{-5}$)}
    \end{subfigure}
    \hfill
    \begin{subfigure}[t]{0.24\linewidth}
        \centering
        \includegraphics[width=\linewidth]{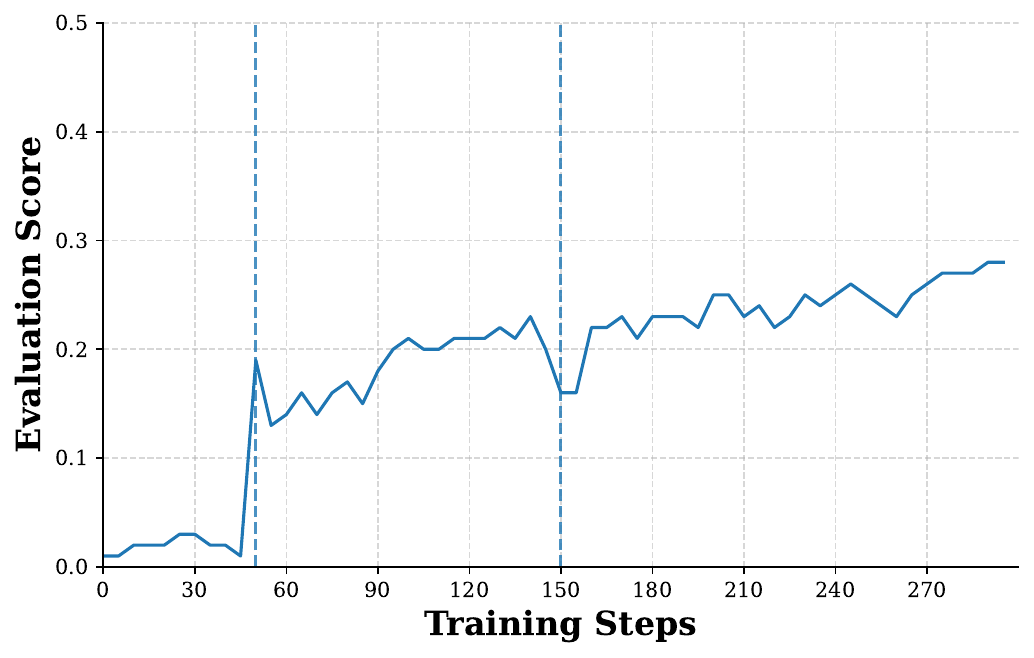}
        \caption{PRS}
    \end{subfigure}

    \caption{Evaluation results of scenario \textit{Miss Param} under different supervisory signals of Qwen2.5-1.5B-Instruct.}
    \label{fig:evaluation_results_miss_param}
\end{figure*}
\begin{figure*}[ht]
    \centering
    \begin{subfigure}[t]{0.24\linewidth}
        \centering
        \includegraphics[width=\linewidth]{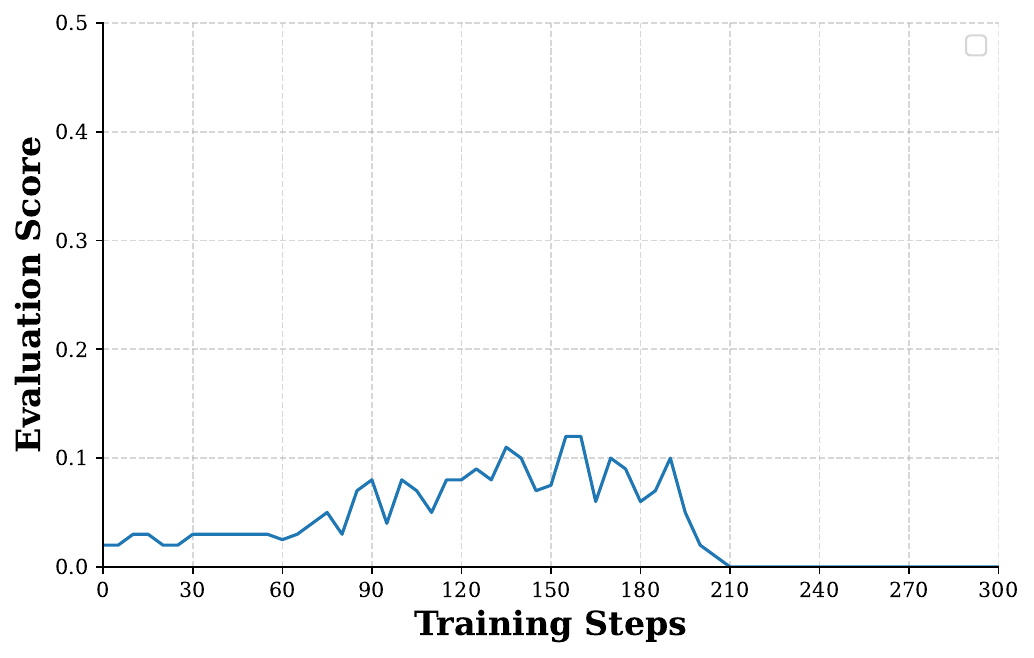}
        \caption{RL}
    \end{subfigure}
    \hfill
    \begin{subfigure}[t]{0.24\linewidth}
        \centering
        \includegraphics[width=\linewidth]{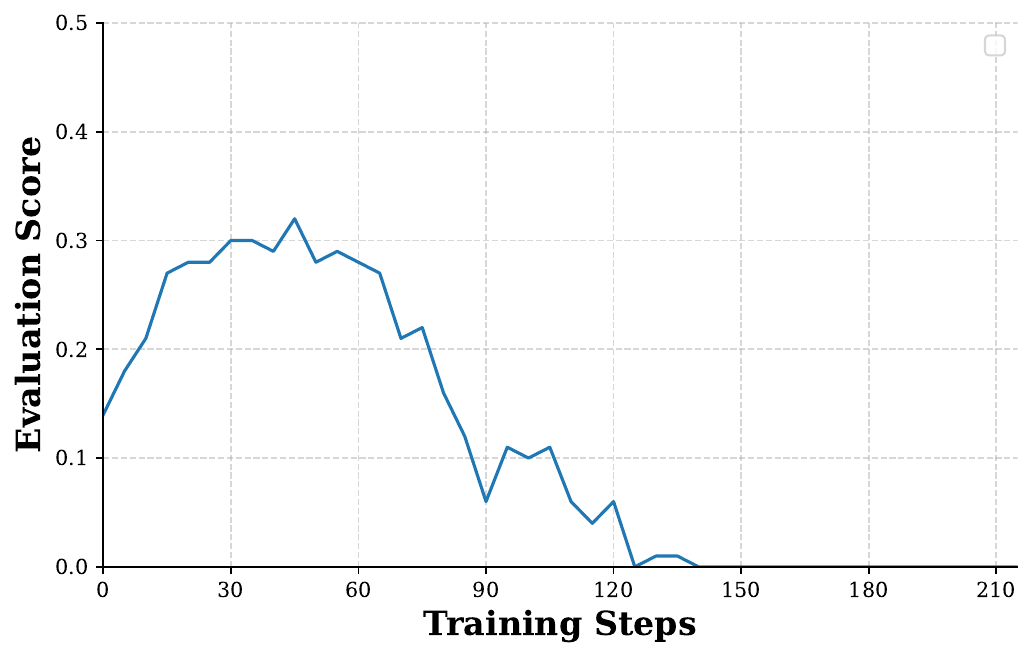}
        \caption{SFT (BFCL) + RL}
    \end{subfigure}
    \hfill
    \begin{subfigure}[t]{0.24\linewidth}
        \centering
        \includegraphics[width=\linewidth]{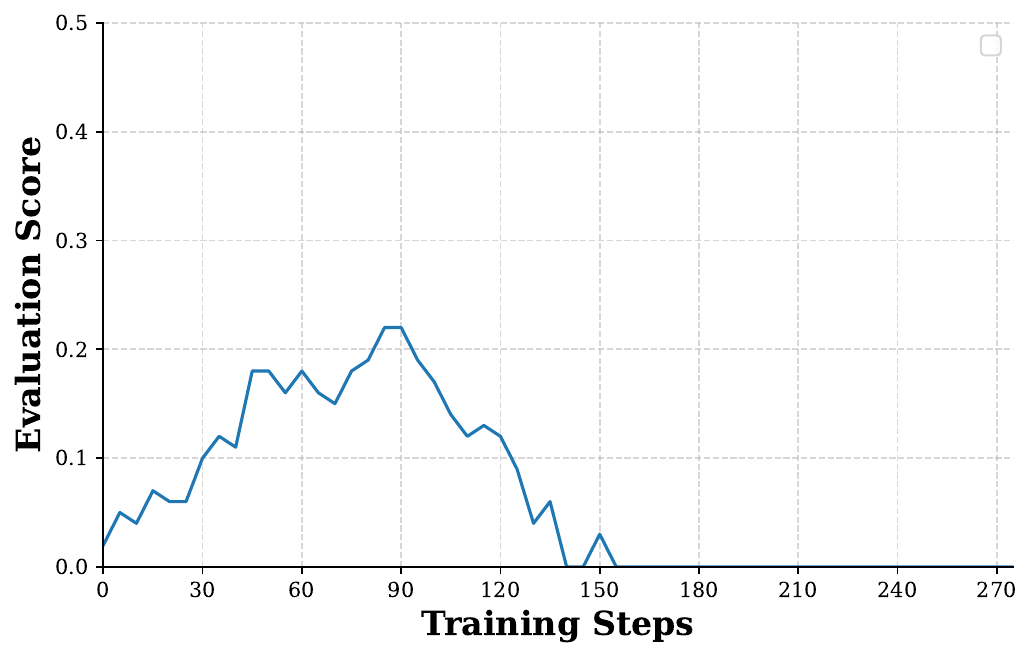}
        \caption{SFT (ToolACE) + RL}
    \end{subfigure}
    \hfill
    \begin{subfigure}[t]{0.24\linewidth}
        \centering
        \includegraphics[width=\linewidth]{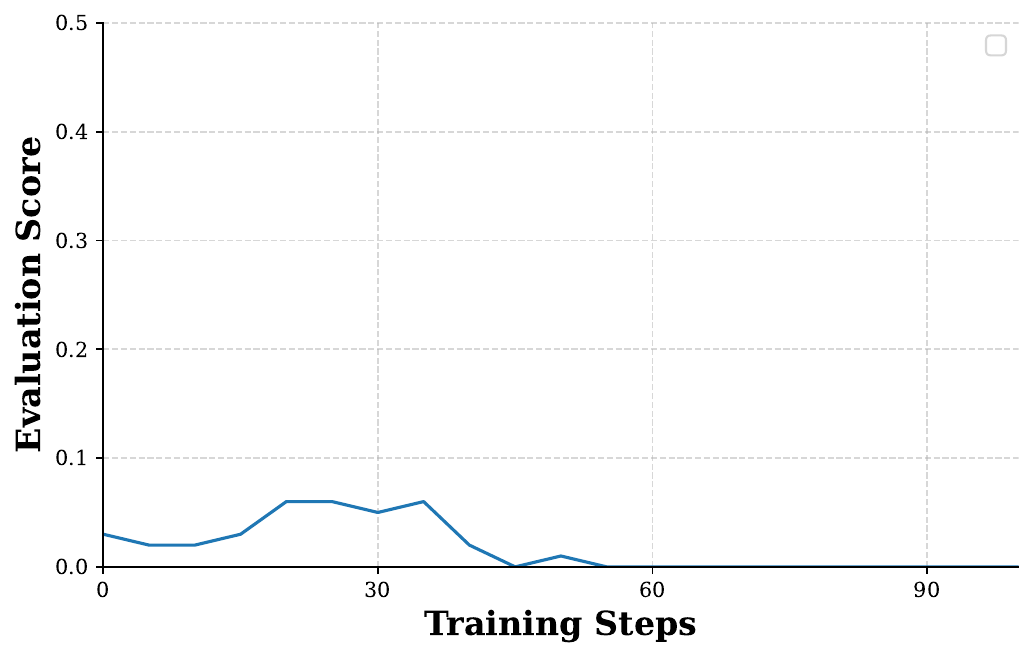}
        \caption{OPS}
    \end{subfigure}

    \begin{subfigure}[t]{0.24\linewidth}
        \centering
        \includegraphics[width=\linewidth]{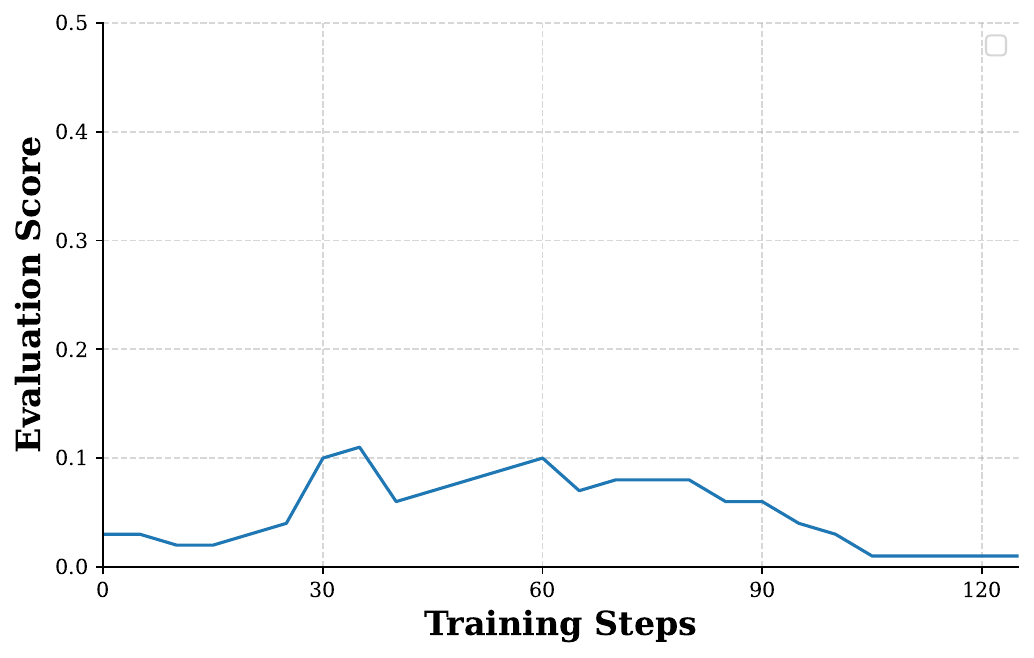}
        \caption{HBG}
    \end{subfigure}
    \hfill
    \begin{subfigure}[t]{0.24\linewidth}
        \centering
        \includegraphics[width=\linewidth]{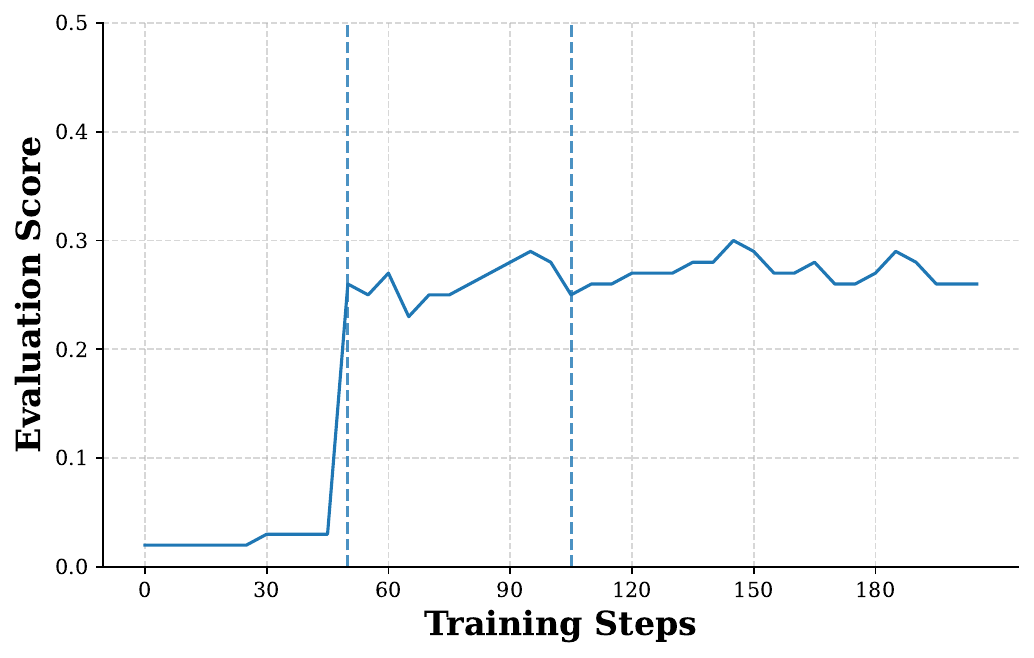}
        \caption{ETS}
    \end{subfigure}
    \hfill
    \begin{subfigure}[t]{0.24\linewidth}
        \centering
        \includegraphics[width=\linewidth]{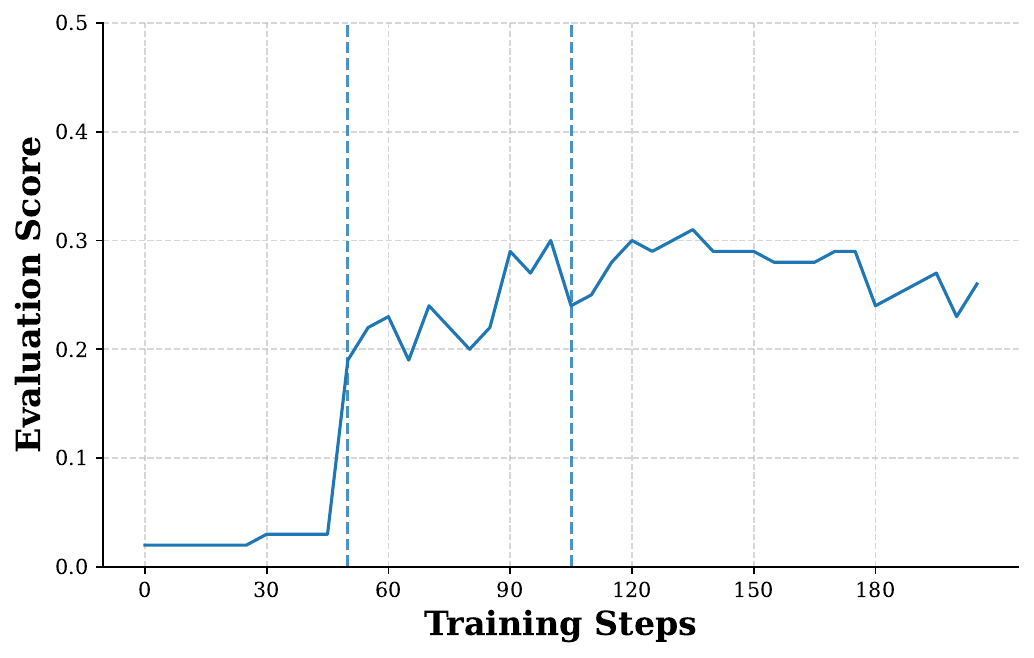}
        \caption{PRS}
    \end{subfigure}

    \caption{Evaluation results of scenario \textit{Base} under different supervisory signals of Qwen3-1.7B.}
    \label{fig:evaluation_results_base3}
\end{figure*}
\begin{figure*}[ht]
    \centering
    \begin{subfigure}[t]{0.24\linewidth}
        \centering
        \includegraphics[width=\linewidth]{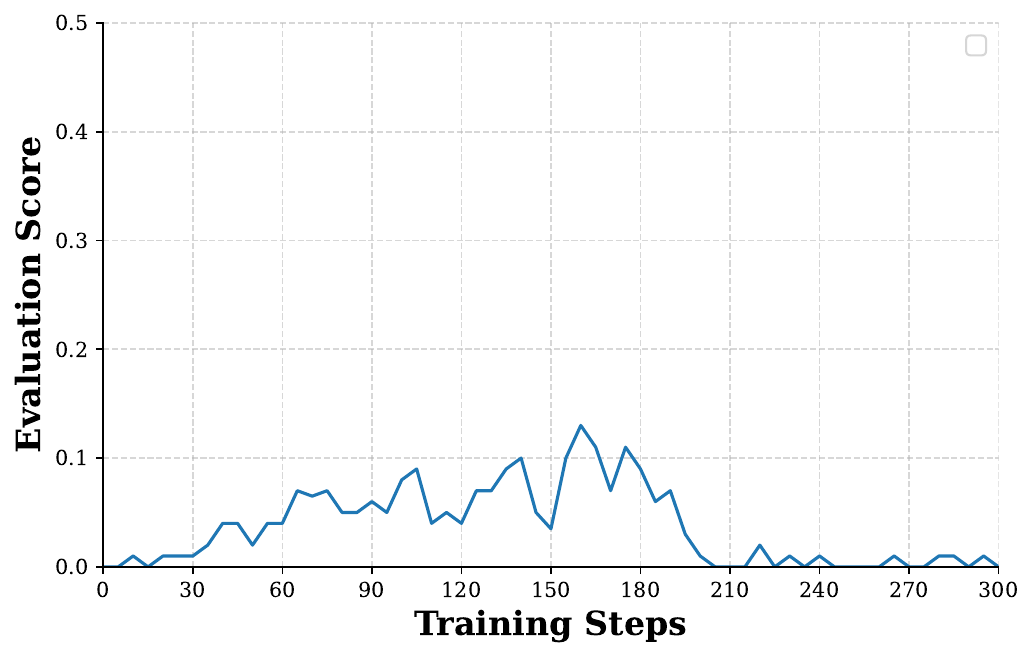}
        \caption{RL}
    \end{subfigure}
    \hfill
    \begin{subfigure}[t]{0.24\linewidth}
        \centering
        \includegraphics[width=\linewidth]{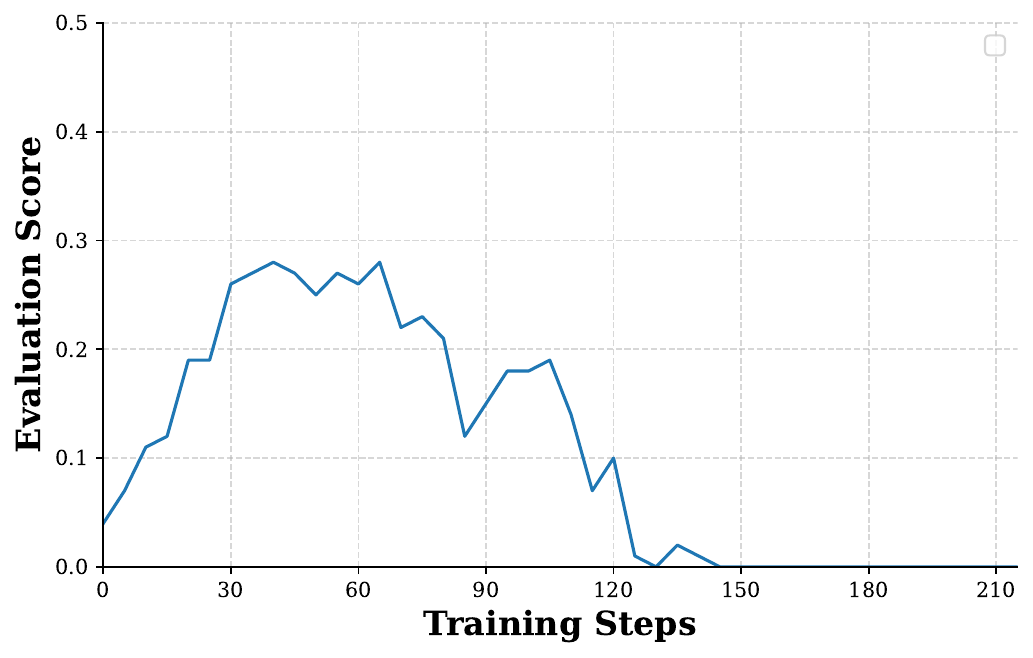}
        \caption{SFT (BFCL) + RL}
    \end{subfigure}
    \hfill
    \begin{subfigure}[t]{0.24\linewidth}
        \centering
        \includegraphics[width=\linewidth]{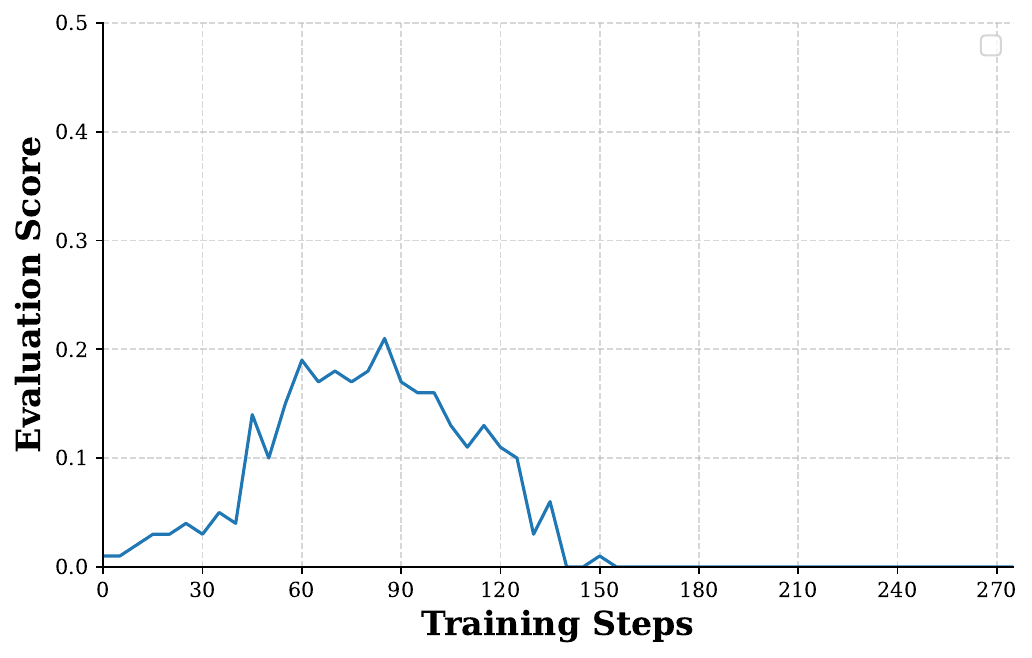}
        \caption{SFT (ToolACE) + RL}
    \end{subfigure}
    \hfill
    \begin{subfigure}[t]{0.24\linewidth}
        \centering
        \includegraphics[width=\linewidth]{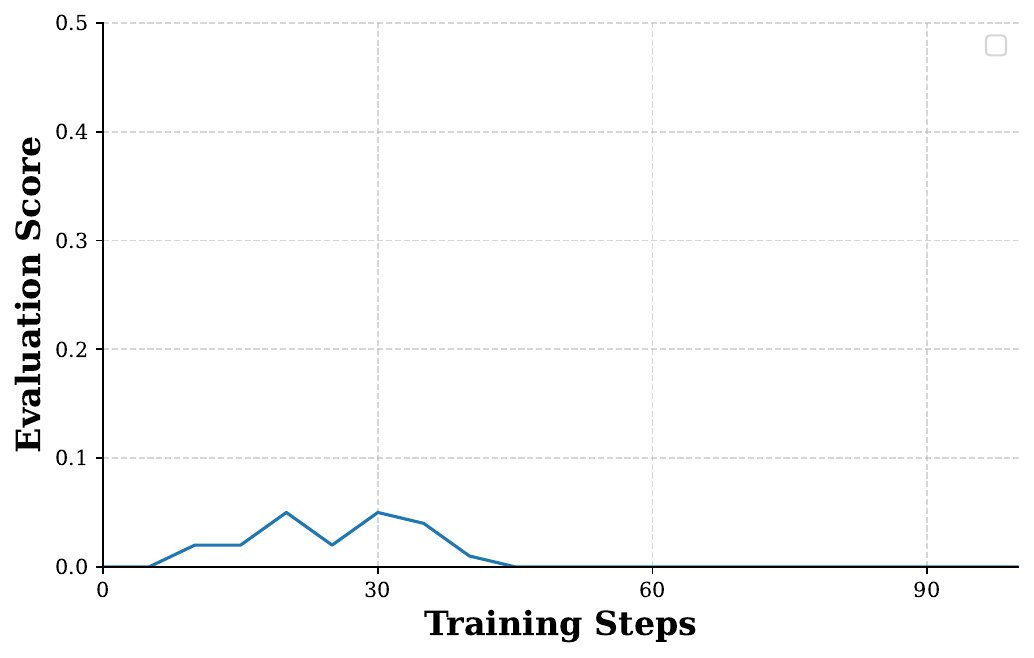}
        \caption{OPS}
    \end{subfigure}

    \begin{subfigure}[t]{0.24\linewidth}
        \centering
        \includegraphics[width=\linewidth]{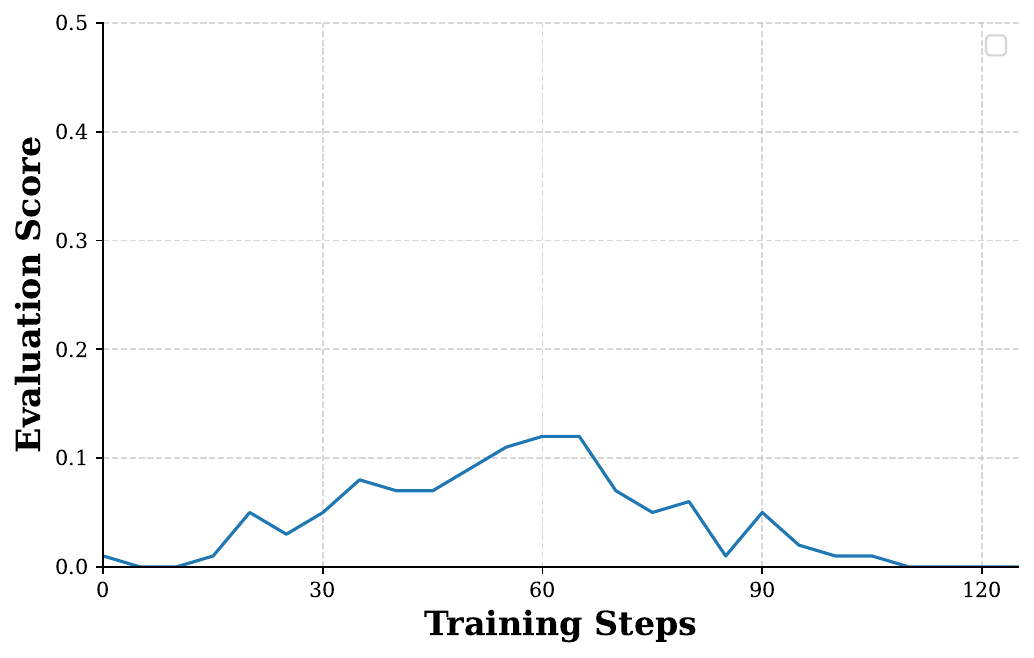}
        \caption{HBG}
    \end{subfigure}
    \hfill
    \begin{subfigure}[t]{0.24\linewidth}
        \centering
        \includegraphics[width=\linewidth]{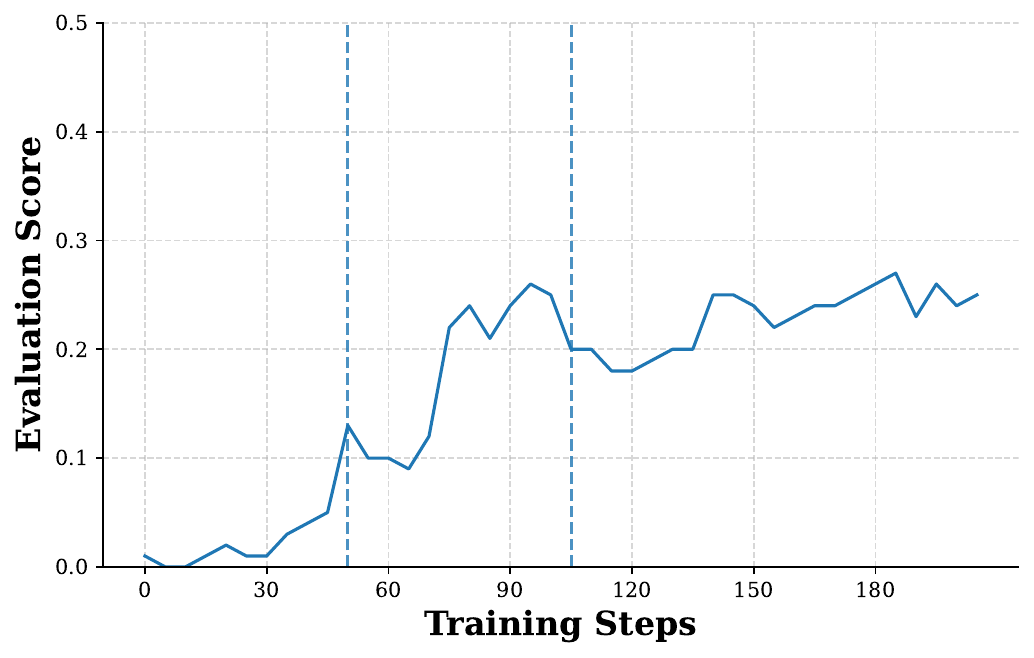}
        \caption{ETS}
    \end{subfigure}
    \hfill
    \begin{subfigure}[t]{0.24\linewidth}
        \centering
        \includegraphics[width=\linewidth]{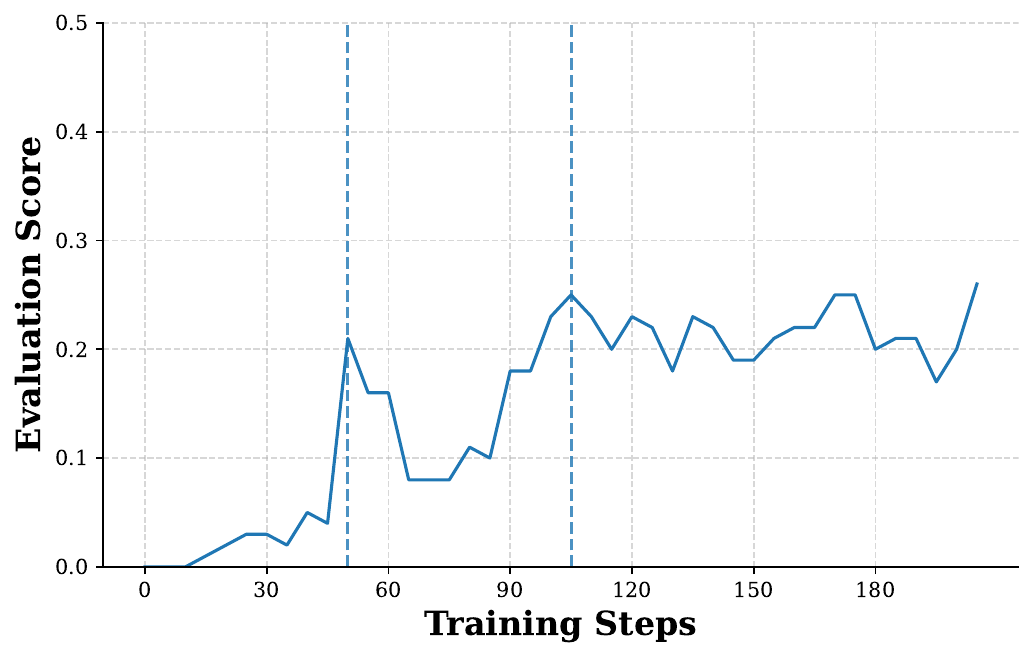}
        \caption{PRS}
    \end{subfigure}

    \caption{Evaluation results of scenario \textit{Miss Func} under different supervisory signals of Qwen3-1.7B.} 
    \label{fig:evaluation_results_miss_func3}
\end{figure*}
\begin{figure*}[ht]
    \centering
    \begin{subfigure}[t]{0.24\linewidth}
        \centering
        \includegraphics[width=\linewidth]{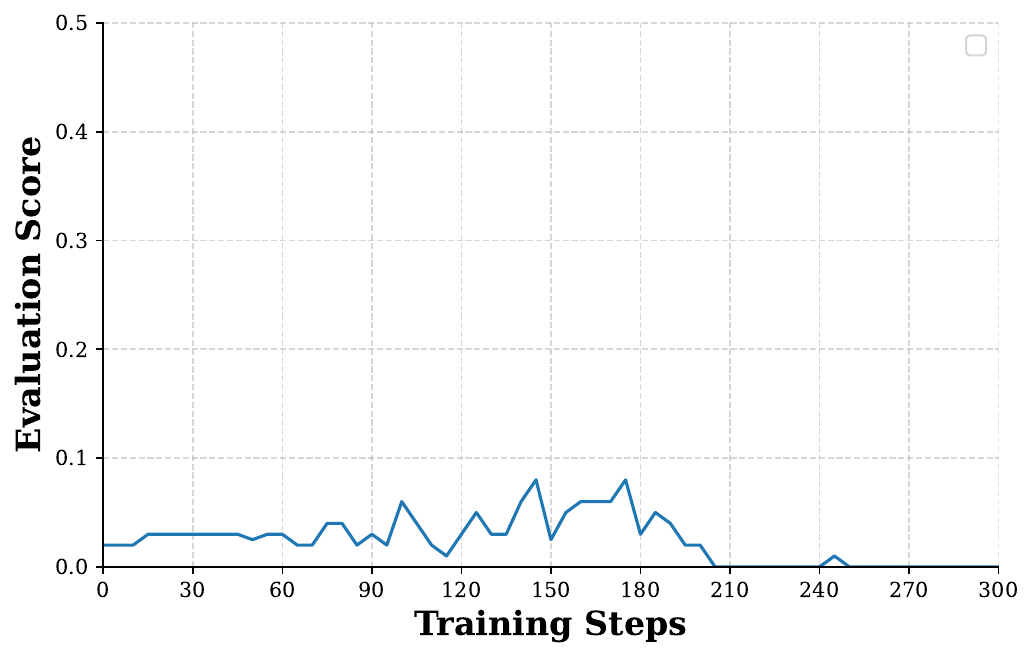}
        \caption{RL}
    \end{subfigure}
    \hfill
    \begin{subfigure}[t]{0.24\linewidth}
        \centering
        \includegraphics[width=\linewidth]{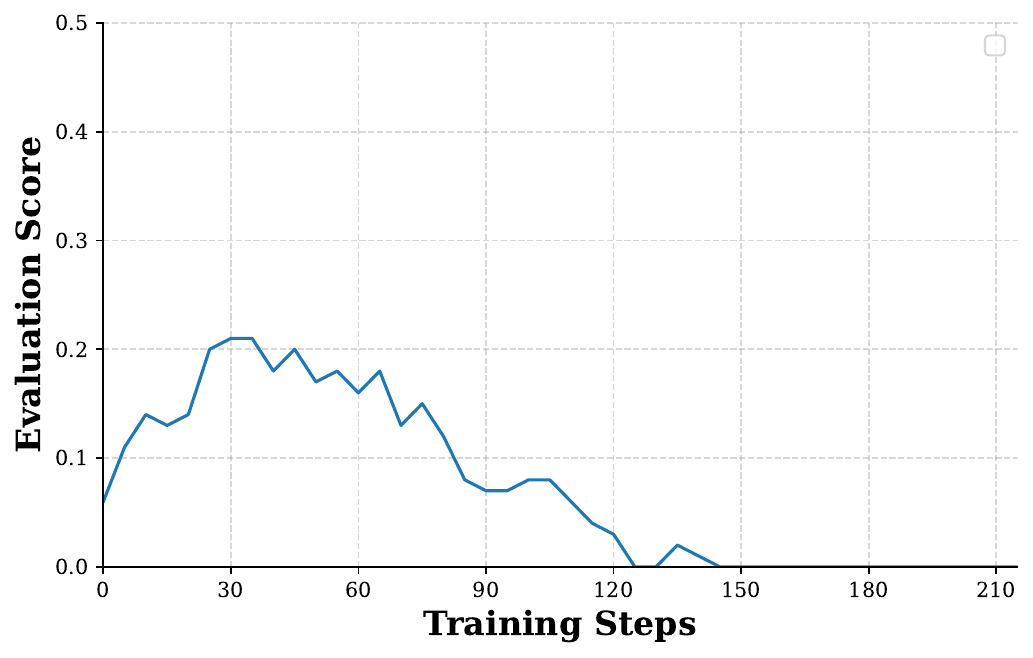}
        \caption{SFT (BFCL) + RL}
    \end{subfigure}
    \hfill
    \begin{subfigure}[t]{0.24\linewidth}
        \centering
        \includegraphics[width=\linewidth]{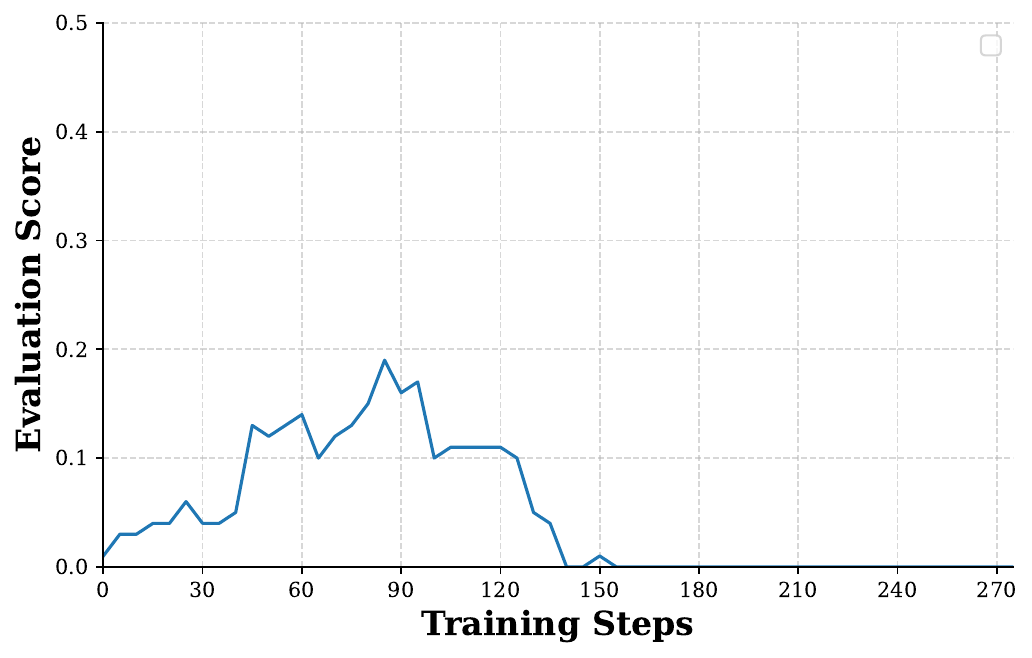}
        \caption{SFT (ToolACE) + RL}
    \end{subfigure}
    \hfill
    \begin{subfigure}[t]{0.24\linewidth}
        \centering
        \includegraphics[width=\linewidth]{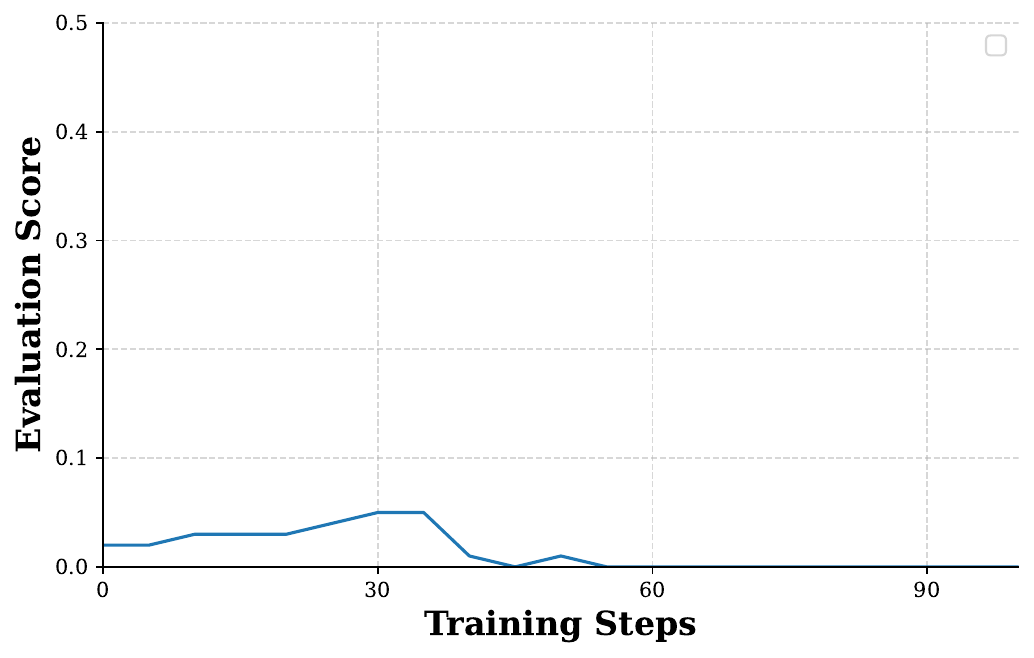}
        \caption{OPS}
    \end{subfigure}

    \begin{subfigure}[t]{0.24\linewidth}
        \centering
        \includegraphics[width=\linewidth]{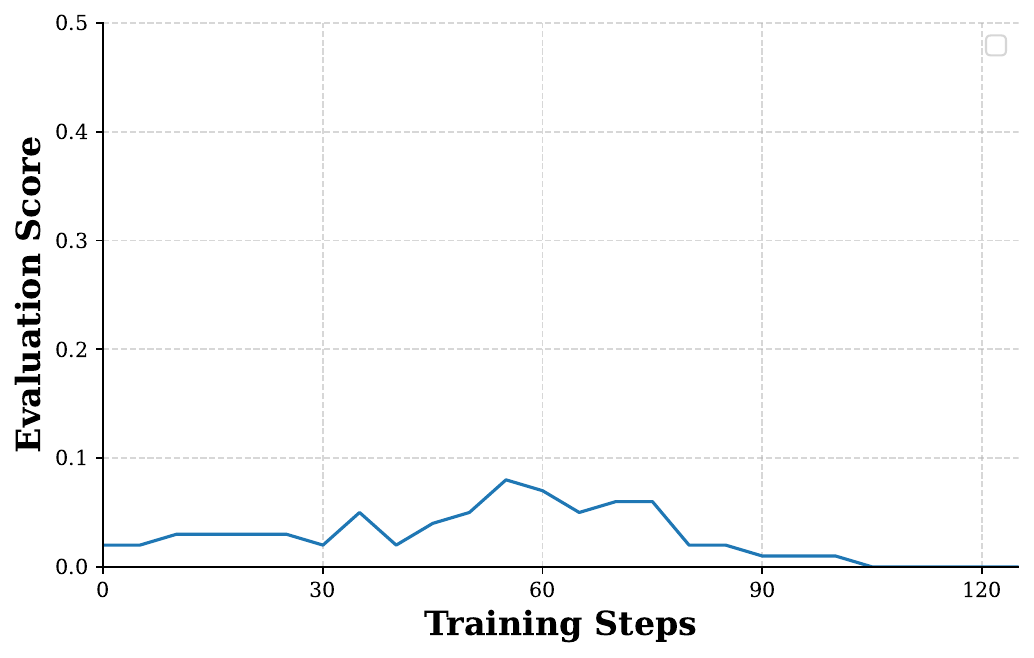}
        \caption{HBG}
    \end{subfigure}
    \hfill
    \begin{subfigure}[t]{0.24\linewidth}
        \centering
        \includegraphics[width=\linewidth]{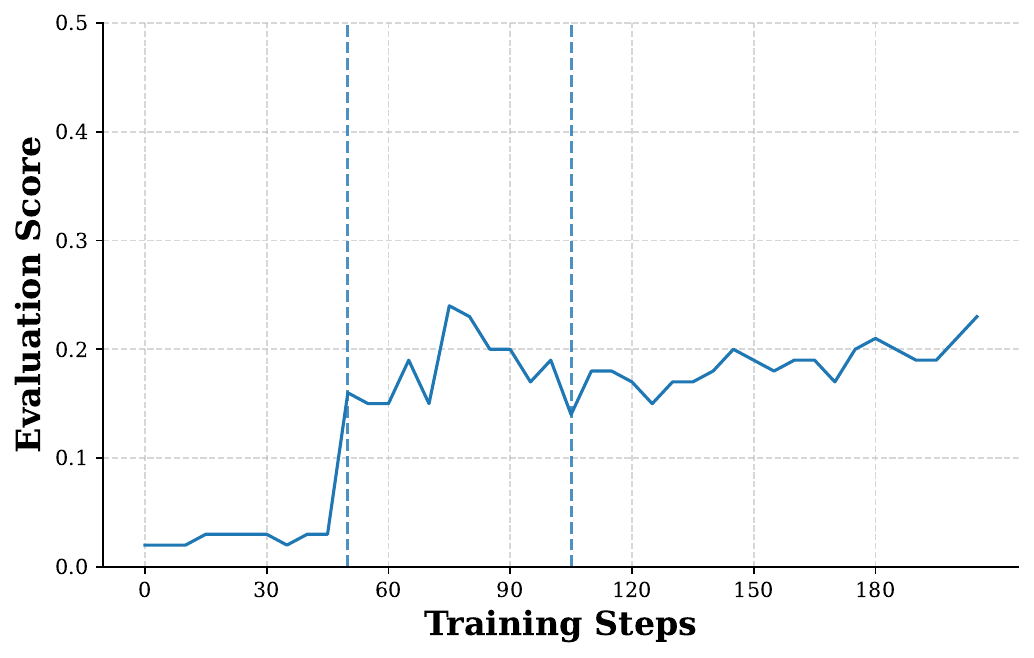}
        \caption{ETS}
    \end{subfigure}
    \hfill
    \begin{subfigure}[t]{0.24\linewidth}
        \centering
        \includegraphics[width=\linewidth]{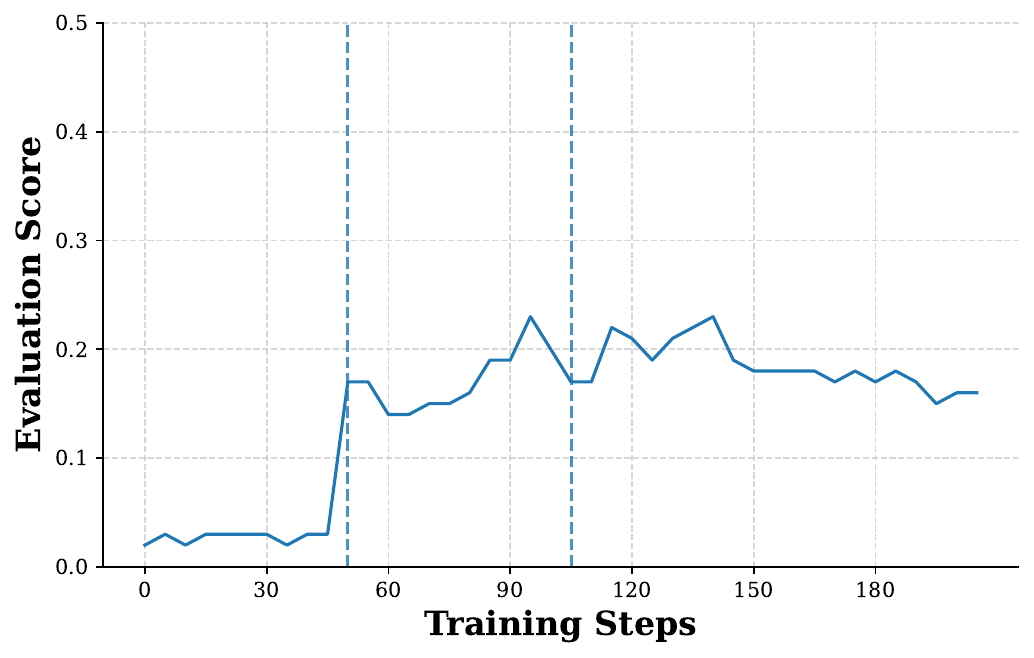}
        \caption{PRS}
    \end{subfigure}

    \caption{Evaluation results of scenario \textit{Miss Param} under different supervisory signals of Qwen3-1.7B.}
    \label{fig:evaluation_results_miss_param3}
\end{figure*}

Since evaluation is conducted using the final checkpoint, some models may demonstrate strong performance during intermediate training stages but fail to retain these gains at convergence. We report the evaluation results in the training process as follows. We can find the best performance during the training process. The results of Qwen2.5 is shown in Figure~\ref{fig:evaluation_results_base}, Figure~\ref{fig:evaluation_results_miss_func} and Figure~\ref{fig:evaluation_results_miss_param}. The results of Qwen3 is shown in Figure~\ref{fig:evaluation_results_base3}, Figure~\ref{fig:evaluation_results_miss_func3} and Figure~\ref{fig:evaluation_results_miss_param3}.

\section{Training Dynamic}
\begin{figure*}[t]
    \centering
    \begin{minipage}[b]{0.48\linewidth}
        \centering
        \includegraphics[width=\linewidth]{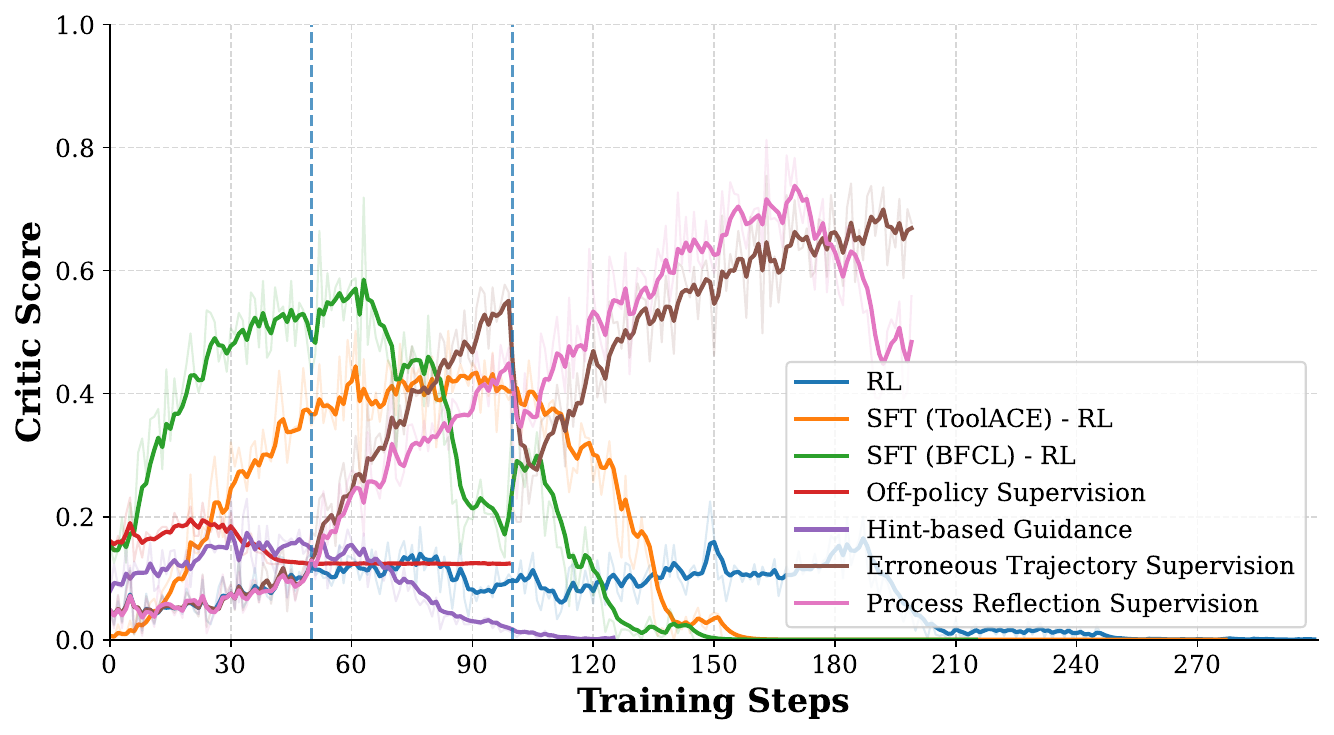}
    \end{minipage}
    \hfill
    \begin{minipage}[b]{0.48\linewidth}
        \centering
        \includegraphics[width=\linewidth]{pic/training_curves_kl.pdf}
    \end{minipage}
    \caption{The training dynamics during the training process on the BFCL-V3 dataset of Qwen3-1.7B.}
    \label{fig:curves_training_con3}
\end{figure*}

The Training dynamics of Qwen2.5 are shown in Figure \ref{fig:curves_training_con} and Qwen3 are shown in Figure \ref{fig:curves_training_con3}.

\newpage
\onecolumn
\FloatBarrier
\section{Prompt and Example}
\label{sec:appendix_prompt_and_example}
The prompt of RPS is in Prompt~\ref{prompt:process_analysis_prompt} and the example is in Example~\ref{prompt:process_analysis_example}.

\begin{tcolorbox}[width=\textwidth, breakable]
{
\renewcommand{\lstlistingname}{Prompt}
\begin{lstlisting}[
    frame=none,
    basicstyle=\ttfamily\footnotesize,
    breaklines=true,
    breakatwhitespace=true,
    breakindent=0pt,
    columns=fullflexible,
    caption={The RPS prompt},
    captionpos=b,
    label={prompt:process_analysis_prompt}
]
You are an expert in intelligent agent tool-use analysis.

Below is the erroneous interaction between a model and tools (including the ground-truth tool-use sequence for reference). Your task is to provide a **comprehensive, layered, and actionable analysis**, and **generalize to similar scenarios** to create augmentation data.  
**Note:** You should write in natural language (not structured JSON), but every “correct tool call” must strictly follow the raw <tool_call> ... </tool_call> format.

Inputs to be filled by caller:
- Available Tools: {available_tools}
- Model Interaction Log (Error Turn): {formatted_summary}

Your required outputs:

1. **Core Error Analysis (must be first)**  
   - Briefly state the root cause of the error (1-2 lines)  
   - Provide 2-4 pieces of evidence from the interaction log to support your conclusion  
   - Provide one immediate and actionable fix recommendation

2. **Generalization: Generate 3-5 similar scenarios**  
   For each scenario, describe in natural paragraphs:
   - One-sentence scenario description (what & why it occurs)  
   - The user request and necessary context  
   - A typical model mistake that might happen  
   - **The correct tool call sequence**  
     (each step must use the raw `<tool_call>\n{{...}}\n</tool_call>` block, containing exactly one JSON object with only `name` and `arguments`)  
   - What reasoning procedures can be borrowed from the current failed case  
   - What aspects differ and require different handling  
   - A brief validation strategy for confirming the fix (auto or manual)

   The scenarios must cover different error categories:
   - Missing parameters  
   - Wrong tool selection  
   - Output mismatch  
   - Timeout/failure  
   - Ambiguous request or multi-intent

3. **Provide one scenario that is similar but requires a different solution**  
   - Explain the similarity (shared reasoning or decision features)  
   - Explain clearly why different tool(s) are required  
   - Provide correct tool calls in the same `<tool_call>` format  
   - Provide rationale

4. **Output style and quality requirements**  
   - Natural language, clear paragraph structure  
   - No JSON schemas, no markdown code fences, no meta-explanations  
   - Tool calls must be valid JSON and executable  
   - Provide 1-2 acceptance criteria per scenario for filtering low-quality synthetic data

5. **Example (format reference only, do not repeat unless needed)**:
<tool_call>
{{"name": "search_database", "arguments": {{"query": "latest model evaluation", "top_k": 5}}}}
</tool_call>

---

Please begin by analyzing the root cause of the provided error turn, then generalize to 3–5 similar scenarios (including at least 1 scenario that requires different handling).

\end{lstlisting}
}
\end{tcolorbox}

\begin{tcolorbox}[width=\textwidth, breakable]
{
\renewcommand{\lstlistingname}{Prompt}
\begin{lstlisting}[
    frame=none,
    basicstyle=\ttfamily\footnotesize,
    breaklines=true,
    breakatwhitespace=true,
    breakindent=0pt,
    columns=fullflexible,
    caption={The example of RPS},
    captionpos=b,
    label={prompt:process_analysis_example}
]
Turn 1
1) User intent understanding, correct behavior, and success points
- Intent: Price-check then book a business-class SF→LA flight for Robert Trenton on 2024-11-25 using travel card card_3487 (access code 1293), and obtain an invoice to verify charges.
- Correct behavior: (a) get flight cost to confirm price, (b) book the flight with provided card and access token, (c) retrieve the invoice for verification.
- Success points: All three tools were called in the correct order. Responses show cost $400, booking_id 3426812 and transaction_id 45451592, and retrieve_invoice returned a consistent invoice with cost and booking details. This aligns with the user’s intent and ground truth.

2) Mistake classification and root cause analysis
- Mistakes: None meaningful. Actions, arguments, and sequence match the ground-truth sequence exactly.
- Root cause: N/A.

3) Improvement plan (corrected reasoning and ideal tool call sequence)
- Reasoning: Confirm price before booking to avoid unexpected charges; then book using provided card and access token; finally retrieve the invoice to verify charges and record transaction details.
- Ideal tool call sequence (matches ground truth and log):
<tool_call>
{"name": "get_flight_cost", "arguments": {"travel_from": "SFO", "travel_to": "LAX", "travel_date": "2024-11-25", "travel_class": "business"}}
</tool_call>
<tool_call>
{"name": "book_flight", "arguments": {"access_token": "1293", "card_id": "card_3487", "travel_date": "2024-11-25", "travel_from": "SFO", "travel_to": "LAX", "travel_class": "business"}}
</tool_call>
<tool_call>
{"name": "retrieve_invoice", "arguments": {"access_token": "1293", "booking_id": "3426812"}}
</tool_call>

- Alignment: This is identical to the ground truth sequence and satisfies the user’s requirement to book and verify the invoice.

Turn 2
1) User intent understanding, correct behavior, and success points
- Intent: Post a tweet on Robert Trenton’s travel account: "Loved my flight journey!" with hashtag #TravelDiaries, then retweet it from his travel account to maximize visibility.
- Correct behavior: Before calling post_tweet/retweet, ensure the Twitter account is authenticated (credentials or an authenticated session) and that the correct account is targeted for posting/retweeting. If credentials or an authenticated account identifier are missing, request them rather than calling tools.

2) Mistake classification (fine-grained) and root cause analysis
- Mistake classification:
  - Missing Parameter / Precondition Check Failure: The model attempted post_tweet without confirming authentication or asking for credentials.
  - Incorrect remediation flow: After failed authentication attempt using wrong credentials, the model tried retweeting (still unauthenticated), producing further errors.
- Root cause:
  - The agent did not treat authentication as a required precondition; it attempted actions without validating credentials or asking for clarifying information (username, which account to use).
  - It also attempted to remediate by guessing credentials ("user"/"pass") rather than requesting valid credentials from the user.

3) Improvement plan (corrected reasoning and ideal tool call sequence)
- Corrected reasoning: When a user requests an action that requires authentication but does not provide credentials or indicate an authenticated session, the agent must pause and request the missing authentication information (or ask the user to confirm which authenticated account to use). No tool calls should be made until valid credentials or an authenticated session are provided.
- Action to take now (at this turn): Ask for the necessary authentication details and account identifier. Do not call any tools until the user supplies credentials or instructs a currently-authenticated account to be used.
- No tool calls should be made at this turn. If the user supplies credentials, the correct sequence (ground truth) would be:
<tool_call>
{"name": "authenticate_twitter", "arguments": {"username": "john", "password": "john1234"}}
</tool_call>
<tool_call>
{"name": "post_tweet", "arguments": {"content": "Loved my flight journey!"}}
</tool_call>
<tool_call>
{"name": "retweet", "arguments": {}}
</tool_call>
- How this aligns with ground truth and logic: Ground truth specifies that because of missing parameters, no tool calls should be made; instead, the agent should request credentials. The three-step sequence above is the correct follow-up once credentials are provided (authenticate → post → retweet).

Turn 3
1) User intent understanding, correct behavior, and success points
- Intent: The user supplied Twitter credentials; authenticate with those credentials, post the tweet, and retweet it.
- Correct behavior: Authenticate using username/password, post the requested tweet, then retweet to amplify it.
- Success points: The agent authenticated successfully (authentication_status: true), posted the tweet ("Loved my flight journey!"), and retweeted successfully. The returned post id (10) and retweet success indicate the actions worked and satisfy the user’s request.

2) Mistake classification and root cause analysis
- Mistakes: Minor inconsistency only:
  - The ground-truth post_tweet call used just content (no explicit tags). In the log the tweet included tags in earlier attempts; the final posted tweet in the log did include tags in earlier step but ground truth shows no tags in the tool call. This is a minor divergence in arguments, not a functional error—user asked for hashtag #TravelDiaries, so including the tag is acceptable.
  - The ground truth retweet call had empty arguments; the agent used tweet_id in the log. Either form is acceptable depending on API design; using tweet_id is explicit and clear.
- Root cause: Small differences in how the API was used (presence/absence of tags field, different retweet parameter shapes). These do not change outcome.

3) Improvement plan (corrected reasoning and ideal tool call sequence)
- Corrected reasoning: Authenticate with supplied credentials; once authenticated, post the tweet exactly as requested (include hashtag #TravelDiaries if the user specified it), capture the returned tweet id, then retweet that tweet to increase visibility. Confirm success and return post and retweet statuses to user.
- Ideal tool call sequence (matches ground truth intent; any of the two retweet argument styles is acceptable depending on API—here we show the ground-truth calls and the explicit tweet-id approach):
<tool_call>
{"name": "authenticate_twitter", "arguments": {"username": "john", "password": "john1234"}}
</tool_call>
<tool_call>
{"name": "post_tweet", "arguments": {"content": "Loved my flight journey!"}}
</tool_call>
<tool_call>
{"name": "retweet", "arguments": {}}
</tool_call>
- If using the explicit tweet id returned by post_tweet (recommended for clarity):
<tool_call>
{"name": "retweet", "arguments": {"tweet_id": 10}}
</tool_call>

- Alignment: This aligns with the ground-truth successful sequence: authenticate → post_tweet → retweet. The explicit tweet_id variant ensures the retweet targets the correct post.

Summary Recommendations (cross-turn)
- Always validate preconditions: check for required credentials or tokens before calling authentication-protected tools. If missing, request them.
- Follow a predictable pattern for transactional flows: estimate cost → confirm booking → retrieve invoice.
- When a user supplies credentials, authenticate then perform actions; do not guess credentials.
- When API shapes differ (e.g., retweet accepting {} vs tweet_id), prefer explicit parameters (tweet_id) if available to avoid ambiguity.
\end{lstlisting}
}
\end{tcolorbox}

\end{document}